\documentclass[journal]{IEEEtai}

\usepackage[colorlinks,urlcolor=blue,linkcolor=blue,citecolor=blue]{hyperref}

\usepackage{color,array}

\usepackage{graphicx}

\usepackage{bm} % For \bm
\usepackage{multirow}%
\usepackage{amsmath,amssymb,amsfonts}%
\usepackage{amsthm}%
\usepackage{mathrsfs}%
\usepackage{xcolor}%
\usepackage{textcomp}%
\usepackage{manyfoot}%
\usepackage{booktabs}%
\usepackage{algorithm}%
\usepackage{algpseudocode}
\usepackage{listings}%
\usepackage{colortbl}
\usepackage{mathrsfs}
\usepackage{subcaption} 
\usepackage{wasysym} % Tao circle
\usepackage{threeparttable}
\usepackage{multirow} 
\usepackage[T1]{fontenc}

\usepackage{hyperref}
\usepackage[pscoord]{eso-pic}

\usepackage{xcolor}

\usepackage{soul}
\sethlcolor{black}
\makeatletter
\newif\if@blind
\@blindfalse %use \@blindfalse on final version
\if@blind \sethlcolor{black}\else
   
\fi
\usepackage{titlesec}
\titlespacing\section{0pt}{5pt plus 4pt minus 2pt}{0pt plus 2pt minus 2pt}
\titlespacing\subsection{0pt}{5pt plus 4pt minus 2pt}{1pt plus 2pt minus 2pt}

\begin{document}

% \SetKwComment{Comment}{/* }{ */}

\title{FedImp: Enhancing Federated Learning Convergence with Impurity-Based Weighting}

\author{Hai Anh Tran\href{https://orcid.org/0000-0002-6215-4879}{\includegraphics[scale=0.13]{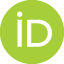}},
       Cuong Ta, 
       Truong X. Tran\href{https://orcid.org/0000-0002-3214-010X}{\includegraphics[scale=0.13]{ORCIDiD_icon64x64.png}}~\IEEEmembership{Senior Member,~IEEE}  % <-this stops a space
% \thanks{Manuscript received 20 February 2025. This paragraph of the first footnote will contain the date on which you submitted your paper for review. It will also contain support information, including sponsor and financial support acknowledgment.}
% \thanks{The next few paragraphs should contain the authors' current affiliations, including current address and e-mail.}
\thanks{Hai-Anh Tran and Cuong Ta are with School of Information and Communication Technology (SOICT), Hanoi University of Science and Technology (HUST), Vietnam. E-mail: anhth@soict.hust.edu.vn, tavietcuong2707@gmail.com}% <-this % stops a space
\thanks{Truong X. Tran is with School of Science, Engineering and Technology, Penn State Harrisburg, The Pennsylvania State University, USA.  Email: truong.tran@psu.edu}% <-this % stops a space
\thanks{Corresponding author: Truong X. Tran; Email: truong.tran@psu.edu}

\thanks{\copyright 2025 IEEE. Personal use of this material is permitted. Permission from IEEE must be obtained for all other uses, in any current or future media, including reprinting/republishing this material for advertising or promotional purposes, creating new collective works, for resale or redistribution to servers or lists, or reuse of any copyrighted component of this work in other works.}}

\markboth{Journal of IEEE Transactions on Artificial Intelligence, Vol. 00, No. 0, September 2025}{}
%{First A. Author \MakeLowercase{\textit{et al.}}: FedImp: Enhancing Federated Learning Convergence with Impurity-Based Weighting}

\maketitle
\IEEEpeerreviewmaketitle
%\advance\dimen0 by -40.75pc\relax

\begin{abstract}
Federated Learning (FL) is a collaborative paradigm that enables multiple devices to train a global model while preserving local data privacy. A major challenge in FL is the non-Independent and Identically Distributed (non-IID) nature of data across devices, which hinders training efficiency and slows convergence. To tackle this, we propose Federated Impurity Weighting (FedImp), a novel algorithm that quantifies each device’s contribution based on the informational content of its local data. These contributions are normalized to compute distinct aggregation weights for the global model update. Extensive experiments on EMNIST and CIFAR-10 datasets show that FedImp significantly improves convergence speed, reducing communication rounds by up to 64.4\%, 27.8\%, and 66.7\% on EMNIST, and 44.2\%, 44\%, and 25.6\% on CIFAR-10 compared to FedAvg, FedProx, and FedAdp, respectively. Under highly imbalanced data distributions, FedImp outperforms all baselines and achieves the highest accuracy. Overall, FedImp offers an effective solution to enhance FL efficiency in non-IID settings.
\end{abstract}

\begin{IEEEImpStatement}
Federated Learning (FL) plays a crucial role in training models without compromising data privacy. This work paves the way for future advancements in adaptive federated learning techniques, ensuring that AI models can be trained more efficiently and equitably across diverse and decentralized data sources. By ensuring faster and more reliable convergence, FedImp enhances the feasibility of FL in large-scale deployments, particularly in resource-constrained environments such as mobile networks, IoT systems, and medical diagnostics. 
\end{IEEEImpStatement}

\begin{IEEEkeywords}
Federated Learning, Fast Convergence Rate, Model Aggregation, Collaborative Machine Learning
\end{IEEEkeywords}

%%\pacs[JEL Classification]{D8, H51}

%%\pacs[MSC Classification]{35A01, 65L10, 65L12, 65L20, 65L70}

\section{Introduction}

Artificial Intelligence and Machine learning power many data-driven applications, but traditional centralized training raises privacy and security concerns. Federated learning (FL) addresses these by enabling decentralized model training across devices without sharing raw data \cite{mammen2021federated}. 

This approach preserves privacy, reduces data transfer, and is ideal for industries with strict confidentiality requirements. The FL process unfolds through a sequence of iterative model updates \cite{kairouz2021advances}. Each local node (client node, device) computes a local model update based on its unique data, capturing local patterns. These updates are then aggregated (e.g., via weighted averaging) into a unified global model, effectively learning from diverse data across all participating devices.

One of the crucial challenges faced by FL algorithms is the presence of non-Independently and Identically Distributed (non-IID) data across decentralized devices \cite{zhu2021federated}. In FL environment, data is often non-IID due to factors like location or user behavior, making it harder to aggregate local models effectively, leading to slow convergence and poor generalization. To mitigate Non-IID data effects, several strategies are employed.  Weighted aggregation allows devices with higher quality or more relevant data to have a greater impact on the global model. Data augmentation artificially expands the diversity of local datasets, improving the model's ability to generalize.  In addition, adaptive learning rates allow the model to adjust its learning process based on the unique characteristics of each device's non-IID data. Several algorithms aggregate local models into a global model for optimal results \cite{liu2020systematic}. Among them, FedAvg \cite{mcmahan2017communication} is widely used, averaging model parameters from participating nodes. However, it often suffers from slow convergence and reduced accuracy in non-IID settings, where node contributions can vary significantly. To improve this, FedAdp \cite{wu2021fast} assigns weights based on the angle between local and global gradients. Yet, when many nodes are non-IID, the global gradient can become misleading, failing to reflect the true descent direction of the global objective. In such cases, local gradients may lead the aggregated gradients to point away from the global minimum, slowing learning or causing instability. This motivates the need for an alternative that better captures the informativeness and diversity of local data, such as the approach used in FedImp.

This research aims to address the highlighted limitations associated with two FL algorithms, FedAvg and FedAdp, by introducing a novel algorithm termed Federated Impurity Weighting (FedImp). Our observation highlights the varying contributions of nodes during global model aggregation. We measure a node’s contribution by assessing the informational impurity in its data using entropy. These contributions are then normalized into distinct weights for aggregation. The proposed strategy enhances generalization across diverse data distributions and mitigates issues where high weights are assigned to nodes with unrepresentative data. FedImp is designed to prevent such biases, ensuring a more balanced and effective global model aggregation in FL.

This paper encompasses the following contributions:

\begin{itemize}
    \item {\textbf{Determining traditional FL algorithm limitations}}: The initial contribution involves a critical examination of the FedAvg and FedAdp algorithms and highlighting their limitations in certain scenarios of non-IID data.
    \item \textbf{Novel Aggregation Strategy}: Introducing FedImp, a new FL algorithm that quantifies the informational richness (impurity) of each client's data using entropy. 
    \item \textbf{Enhanced FL Convergence and Accuracy}: By addressing the challenges posed by non-IID data distributions, the proposed method can improve the convergence rate and overall accuracy of the global model.
\end{itemize}

We have implemented the proposed algorithm and conducted experiments to evaluate its performance under various FL scenarios using the EMNIST and CIFAR-10 datasets. The results show that FedImp achieves significantly faster convergence than FedAvg \cite{liu2020systematic}, FedProx \cite{li2020federated}, and FedAdp \cite{mcmahan2017communication} in FL with non-IID data. Experiments show reductions in communication rounds by up to 64.4\%, 27.8\%, and 66.7\% compared to FedAvg, FedProx, and FedAdp, respectively, on the EMNIST dataset. For the tests with the CIFAR-10 dataset, FedImp can also reduce the number of communication rounds, ranging up to 44.2\%, 44\%, and 25.6\% compared to FedAvg, FedProx, and FedAdp, respectively. In the experiment with highly local imbalanced data nodes, FedImp is particularly advantageous and often becomes the only algorithm to reach target accuracy.

The rest of this paper is organized as follows: Section \ref{sec:related_work} discusses the related works. Section \ref{sec:preliminary} provides the foundation of FL and algorithms: FedAvg, FedProx, and FedAdp. In Section \ref{sec:methodology}, the proposed algorithm is presented. Experimental implementation and results are shown in Section \ref{sec:experiment}, and the conclusion is presented in Section \ref{sec:conclusion}.

\section{Related Work}
\label{sec:related_work}
Numerous methods have been suggested for consolidating local models into a single global model in order to attain desirable outcomes. McMahan et al. \cite{mcmahan2017communication} introduced the Federated Averaging (FedAvg) algorithm, in which clients perform multiple epochs of SGD on their local datasets and send their models to the server, which averages them to form a new global model. However, the presence of non-IID data at each client can negatively affect the performance of the FedAvg algorithm, including both a slow convergence rate and poor accuracy. Addressing this challenge is crucial for improving the effectiveness of the FL approach, and various solutions have been proposed. Li et al. \cite{li2020federated} introduce FedProx, an extension of FedAvg designed to handle heterogeneity in federated networks. Though its modifications are minor, they have significant effects. Experiments show that FedProx achieves better convergence than FedAvg across real-world datasets, with a 22\% average improvement in test accuracy, especially in highly heterogeneous environments.

SGD with momentum has demonstrated excellent success in speeding up network training in a centralized machine learning approach by accumulating the gradient history over time in order to dampen oscillations. Utilizing this idea, Hsu et al. \cite{hsu2019measuring} proposed the Federated Averaging with Server Momentum (FedAvgM) algorithm. This is especially suitable for FL, where the participating parties may only hold a small subset of labels and a sparse distribution of data. Experiments on CIFAR-10 demonstrate improved classification performance for FedAvgM over FedAvg over a range of non-identicalness, with classification accuracy improved from 30.1\% to 76.9\% in the most skewed settings.

Besides, Yeganeh et al. \cite{yeganeh2020inverse} proposed IDA (Inverse Distance Aggregation), a novel adaptive weighting approach for clients based on meta-information, which handles unbalanced and non-iid data. The IDA method uses model parameter distances to minimize outlier effects and improve convergence. Results show that IDA outperforms FedAvg in classification accuracy in non-IID scenarios and is resilient to low-quality or harmful data from client nodes. Unlike FedAvg, which assumes clients with more data have better distributions, IDA allows aligned clients to exclude out-of-distribution models.

Minimizing the local loss function doesn’t guarantee minimizing the global loss. To address this in non-IID data, Acar et al. introduced FedDyn \cite{acar2021federated}, which dynamically regularizes each client’s loss to align with the global loss. This makes FedDyn robust to varying heterogeneity levels, allowing full minimization per client. It achieves a convergence rate of $O(\frac{1}{T})$ in convex and non-convex settings and a linear rate in strongly convex settings, remaining agnostic to device heterogeneity and resilient to large-scale, unbalanced, and partially participating devices.
 
Realizing the personalization of the global model becomes crucial in handling the challenges that arise with non-IID data, Vahidian et al. \cite{vahidian2021personalized} introduced Personalized FL by Pruning (Sub-FedAvg). Sub-FedAvg enhances efficiency by identifying a small subnetwork per client using hybrid pruning (structured and unstructured). Instead of averaging all parameters like FedAvg, it averages only the remaining parameters in each client's subnetwork. 

In scenarios involving non-IID data, the contributions of participating nodes to the training process are unequal. To address this, Wu et al. introduced the Federated Adaptive Weighting (FedAdp) algorithm in their work \cite{wu2021fast}. This algorithm assigns dynamic weights to update the global model based on each node's contribution per training round. The authors have shown that FL with FedAdp can reduce the number of communication rounds by up to 54.1\% on the MNIST dataset and up to 45.4\% on the FashionMNIST dataset, as compared to the FedAvg algorithm.

While FedAdp offers advancements, it struggles with global gradient misalignment in non-IID settings. When nodes lack label diversity, local gradients deviate, leading to suboptimal global updates. Overweighting nodes with similar gradients can further degrade performance. 

In recent years, privacy-preserving techniques have been extensively explored to enhance the security of FL. Fotohi et al. \cite{fotohi2024lightweight} propose a lightweight framework using differential privacy to protect client updates from adversarial reconstruction. Blockchain integration is another approach that provides decentralized trust and security. A blockchain-enabled FL model can use smart contracts and cryptographic verification to prevent adversarial tampering \cite{fotohi2024decentralized}. While these privacy-oriented methods are beyond the primary scope of FedImp, which focuses on convergence under non-IID conditions, we mention them here to highlight the broader context and ongoing advancements in FL security. We will further discuss potential privacy considerations and future enhancements of our proposed method in the discussion section of the experimental results.

In summary, recent FL research has seen substantial progress through adaptive weighting mechanisms (e.g., FedAdp, IDA), personalization techniques (e.g., Sub-FedAvg), and regularization-based convergence strategies (e.g., FedDyn). While these methods each address aspects of heterogeneity, they often rely on gradient similarity, parameter distance, or client-specific architectures. In contrast, FedImp introduces a fundamentally different perspective by quantifying the informational richness of local datasets through entropy, allowing data-driven weighting that is independent of model-specific signals. This positions FedImp as a general and lightweight approach to improve convergence in non-IID settings without requiring modifications to model structure or optimization methods.

\section{Preliminaries of Federated Learning Algorithms}
\label{sec:preliminary}

This section introduces the fundamental concepts of FL and discusses the principles of the FedAvg, FedProx, and FedAdp algorithms. Additionally, it examines the limitations of FedAdp and FedAvg, providing the motivation for proposing a more comprehensive algorithm.

\subsection{Overview of Federated Learning Algorithms}
FL is a distributed learning approach where clients train a shared model without sharing raw data. It aims to optimize a global loss function, $F(\mathbf{w})$, which aggregates individual client losses in multiple communication rounds. At each communication round \(t\), a subset of \(K\) nodes is selected, and the global model \(\mathbf{w}(t-1)\) from the previous communication round is transmitted to the chosen nodes. Subsequently, each participating node \(i\) executes stochastic gradient descent (SGD) training to minimize its local loss \(F_i(\mathbf{w})\):

\begin{equation}
    \mathbf{w}_i(t) = \mathbf{w}(t-1) - \eta \nabla F_i(\mathbf{w}(t-1))
\end{equation}
where \(\eta\) denotes the learning rate, and \(\nabla F_i(\mathbf{w}(t-1))\) represents the gradient at node \(i\). FL algorithms then seek to update the global model parameters \(\mathbf{w}\) by aggregating the local model updates (either parameters or gradients) from each node:

\begin{equation}
    \label{eq:update_weight}
    \mathbf{w}(t)=\sum^{K}_{i=1}\psi_i\mathbf{w}_i(t)
\end{equation}
Here, \(\psi_i\) is the weighting factor of node \(i\) when calculating the global parameters. Different algorithms use various weighting strategies. The following part explains node weight determination for FedAvg and FedAdp and the loss function modification in FedProx.

\textit{Federated Averaging (FedAvg) Algorithm}: 
The FedAvg \cite{mcmahan2017communication} algorithm aggregates local model updates from participating nodes based on their dataset sizes. While it is often described as treating node contributions equally in terms of participation, the actual aggregation weights are proportional to the number of training samples each client possesses. Specifically, the weighting factor of each node in the global aggregation is given by:
\begin{equation}
\psi_i(t) = \frac{D_i}{\sum_{i'=1}^{K} D_{i'}}
\end{equation}
where \( D_i \) is the number of local training samples on client \( i \). When all \( D_i \) are identical, this reduces to equal weighting across clients.

%\subsection{Background of FedProx Algorithm}
\textit{FedProx Algorithm}: %To address the issue of statistical heterogeneity across participating nodes when minimizing the local loss function may not equate to minimizing the global loss function, Li et al. \cite{li2020federated} propose to add a proximal term to the local subproblem, which restricts the local updates to be closer to the global model. 
To tackle statistical heterogeneity, Li et al. \cite{li2020federated} introduce a proximal term in the local subproblem, keeping local updates closer to the global model. In particular, instead of just minimizing the local function \(F_i(\mathbf{w})\), node \(i\) minimizes the following objective \(h_i\):
\begin{equation}
    \min_{\mathbf{w}} h_i(\mathbf{w}) = F_i(\mathbf{w}) + \frac{\mu}{2} \|\mathbf{w} - \mathbf{w}(t-1)\|^2
\end{equation}
where \(\mathbf{w}(t-1)\) is the global parameters at previous rounds and \(\mu\) is the weight of the proximal term used. When \(\mu\) equals 0, this strategy is equivalent to FedAvg. A small \(\mu\) may not make any difference when compared to FedAvg, while a large \(\mu\) may potentially slow the convergence by forcing the local parameters to be close to the global parameters.

%\subsection{Background of Federated Adaptive Weighting (FedAdp) Algorithm}
\textit{Federated Adaptive Weighting (FedAdp) Algorithm}: The FedAdp algorithm \cite{wu2021fast} dynamically assigns weights based on the correlation between local and global gradients. Specifically, it measures node contributions using the angle between local and global gradient vectors and applies a non-linear mapping function to determine their weighting. This approach can make a more adaptive and effective global model update.

\begin{figure}[t!]
  \centering
  \subfloat{\includegraphics[width=0.7\linewidth]{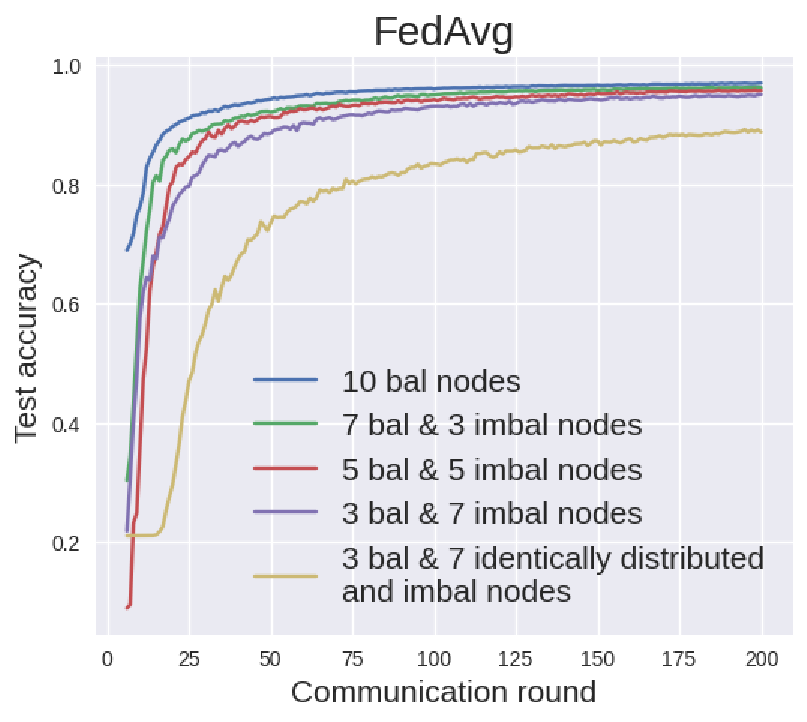}}
  \hfill
  \subfloat{\includegraphics[width=0.7\linewidth]{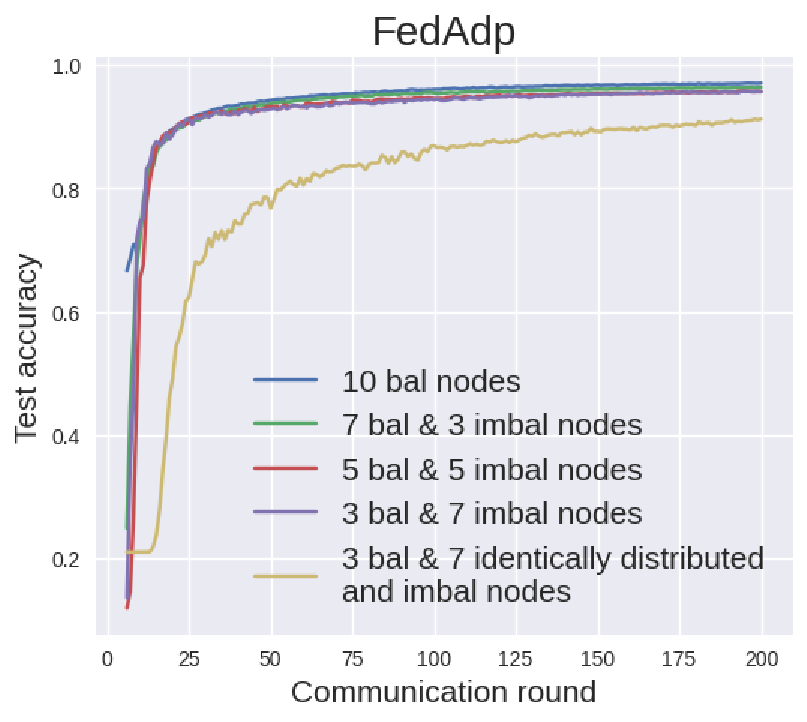}}
  \caption{Illustration of the preliminary test accuracy of FedAvg and FedAdp over communication rounds under different non-IID cases. Demonstrating these algorithms' limitations on identical, imbalanced, and highly skewed data scenarios.}
  \label{fig:mnist}
  \vspace{-10pt}
\end{figure}

\begin{figure*}[t!]
    \centering
    \includegraphics[width=0.85\linewidth]{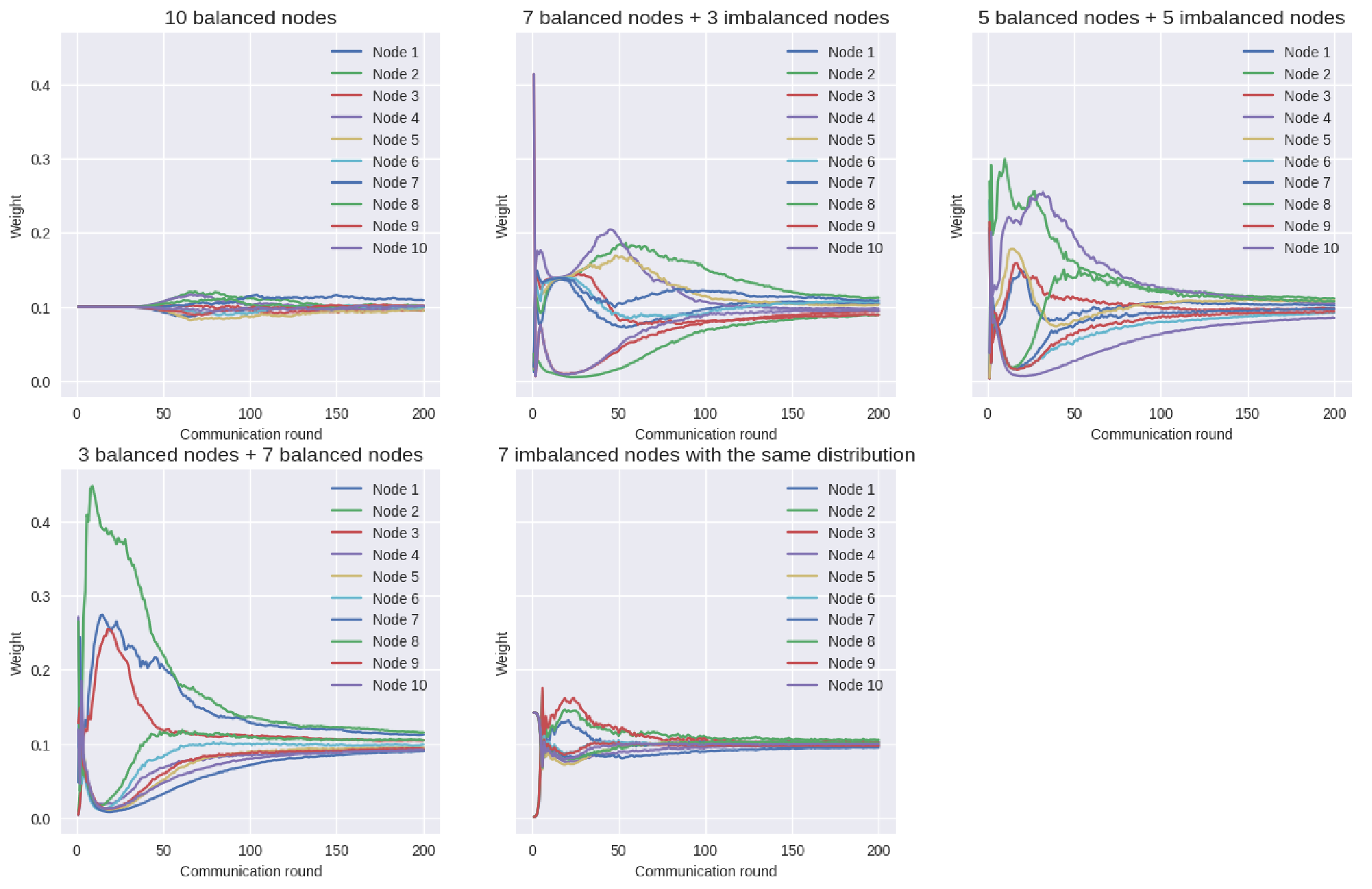}
    \caption{Illustration of FedAdp weighting factor distribution across clients showcasing the evidence that extreme and evenly distributed imbalance reduces weight discrimination, potentially impacting performance.}
    %Weight assigned by the FedAdp algorithm to individual clients over communication rounds.}
    \label{fig:weight}
    \vspace{-10pt}
\end{figure*}

\subsection{Limitation of FedAvg and FedAdp}
The limitation of FedAvg is that it has low accuracy and slow convergence on non-IID data. FedAdp faces issues when the global gradient is misaligned with the expected direction, often due to nodes lacking full label representation. When many nodes exhibit this, averaging their local gradients can lead to suboptimal performance, especially if high weights are assigned to misaligned nodes. 

To further explore these limitations, we re-implement experiments from Wu et al. \cite{wu2021fast}, analyzing FedAvg and FedAdp convergence on a ten-agent CNN model with MNIST under varying non-IID distributions. We also introduce a specific non-IID scenario where FedAdp is expected to underperform.

Figure \ref{fig:mnist} shows the test accuracy over communication rounds of FedAvg and FedAdp on various non-IID data scenarios. FedAvg performs well when there are ten balanced data local nodes, but its performance gradually deteriorates with an increasing number of imbalanced data nodes. On the other hand, FedAdp's performance remains quite stable. However, in 7 imbalanced data nodes with identical data distributions, FedAdp's performance experiences a sudden decline. 

Figure \ref{fig:weight} illustrates the weighting factors assigned by FedAdp over communication rounds. With 10 nodes and balanced data, weights remain stable at around 0.1. In the case of 3 imbalanced nodes, each with samples from only 2 classes, FedAdp assigns higher weights to balanced nodes and lower weights to imbalanced ones. The gap is wider in early rounds but narrows over time. A similar trend appears with 5 and 7 imbalanced nodes, where balanced nodes receive higher weights. However, with 7 imbalanced nodes of identical distributions, FedAdp performs poorly as the weight differences between diverse and less diverse nodes diminish. Although it might seem intuitive that identically distributed clients should receive equal weights, this becomes problematic when those distributions are heavily skewed. In such cases, FedAdp assigns similar weights to a majority of clients that possess highly imbalanced and non-representative data (e.g., only a small subset of classes), which collectively dominate the aggregation process. This uniform weighting over biased nodes reinforces local data skew and leads the global model to overfit to those limited class patterns, ultimately hindering generalization and slowing convergence.

Overall, FedAvg exhibits suboptimal performance in non-IID scenarios. While FedAdp successfully addresses FedAvg's limitations in many instances, it may still encounter challenges in more complex distribution scenarios.

\section{Methodology: Federated Impurity Weighting}
\label{sec:methodology}
This section describes our methodology for improving FL convergence and performance. We introduce a method for quantifying and weighting each node's data contribution and then present our FedImp algorithm. 

\subsection{Impurity Weighting Updating Factor}
\label{sec:formula}
We use entropy to represent the informational richness of each node's data. Entropy \( S_i \) quantifies the impurity or information content within node \( i \)'s dataset and is defined as:

\begin{equation}
\label{eq:entropy}
S_i = -\sum_{j=1}^{C} p_j \log_C p_j
\end{equation}
where \( C \) denotes the total number of class labels in the federated learning system, and \( p_j \) represents the proportion of samples belonging to class \( j \) at node \( i \). \( p_j \) is computed as:

\begin{equation}
p_j = \frac{|D_{ij}|}{|D_i|}
\end{equation}
where \( |D_{ij}| \) is the number of local samples at node \( i \) that belong to class \( j \), and \( |D_i| \) is the total number of training samples at node \( i \). This definition ensures that the entropy score \( S_i \) is computed consistently across nodes with varying local class distributions. Note that \( \log_C \) denotes the logarithm to the base \( C \), which ensures that \( S_i \) is normalized between 0 and 1. Higher entropy reflects greater data diversity or uncertainty, while lower entropy indicates more homogeneous or certain data.

The value of \(S_i\) lies within the inclusive interval from 0 to 1. The closer \(S_i\) is to 0, the more concentrated the dataset is around a single class, while the closer \(S_i\) is to 1, the more evenly distributed the dataset is among different classes, indicating higher uncertainty or impurity. Expressly, \(S_i\) assumes a value of 0 when the examples within the dataset exclusively pertain to a single class, reflecting a state of utmost certainty. On the contrary, \(S_i\) attains a value of 1 when the distribution of examples across various classes is perfectly equal, illustrating a scenario of maximum uncertainty.

\textbf{Weighting Factor}: To weight nodes during global model aggregation, we use a Softmax function (with temperature \(\tau>0\)) applied to the impurity values calculated for each node:

\begin{equation}
\label{eq:weight}
    \psi_i(t)=\frac{D_i e^{\frac{S_i}{\tau}}}{\sum^{K}_{i'=1} D_{i'} e^{\frac{S_{i'}}{\tau}}} 
\end{equation}
where:  
\begin{itemize}
    \item \( \psi_i(t) \) represents the aggregation weight assigned to client \( i \) at communication round \( t \).
    \item \( D_i \) denotes the number of local training data at client \( i \).
    \item \( S_i \) is the entropy-based impurity score of client \( i \), computed in equation (\ref{eq:entropy}).
    \item \( \tau \) is the temperature parameter, which controls the sharpness of the weighting distribution. A lower \( \tau \) increases the contrast between high- and low-entropy clients.
    \item \( K \) is the number of clients participated in each round.
\end{itemize}

Choosing an appropriate value for the temperature parameter \( \tau \) is crucial. If \( \tau \) is set too low, the softmax function becomes sharp, resulting in overly aggressive weighting that may exaggerate small differences in entropy and disproportionately favor a few nodes. Conversely, a high value of \( \tau \) leads to flatter weighting, making the aggregation resemble uniform averaging and diminishing the benefit of impurity-aware differentiation. A suitable \( \tau \) balances these effects to guide meaningful contributions from diverse clients while avoiding dominance or dilution. A detailed empirical study on the sensitivity of \( \tau \) is provided later in Section~\ref{sec:experiment}.

After forming weights for the participating nodes, the global model parameters \(\mathbf{w}\) are updated as in equation (\ref{eq:update_weight}).

The temperature parameter \(\tau\) serves as a parameter controlling the difference between the weights of nodes. A small \(\tau\) emphasizes the contribution of nodes with more information while diminishing the influence of nodes with less information.
However, it is crucial to note that using a small value of \(\tau\) is not universally effective. In a hypothetical scenario where nodes with less information possess samples belonging to classes not present in nodes with more information, it would not be reasonable to diminish the influence of the less informative nodes significantly. In such cases, maintaining a certain level of influence from these nodes becomes essential, as they contribute unique information that is not available in nodes with more comprehensive data. Striking a balance in the FL process is crucial to ensure that all relevant information is considered and integrated, even from nodes with less data, to avoid losing valuable insights or potential improvements for the entire model. The subsequent section empirically verifies the effect of different \(\tau\) values.

\subsection{FedImp Algorithm}
\label{sec:alg}

\begin{algorithm}[h!]
    \caption{Federated Impurity Weighting Algorithm}
    \label{alg:fedimp}
    \begin{algorithmic}[1]
        \Require\(T\): the number of communication rounds, \(K\): the number of participating nodes in each round, \(B\): the local minibatch size, \(E\): the number of local epochs, \(\eta\): the learning rate, \(\tau\): the control parameter
        
        \State \textbf{Server execute}:
        \State Initialize \(\mathbf{w(0)}\).
        \For{\(t = 1, 2, \ldots, T\)}
            \State Choose random set of \(K\) nodes.
            \For{each node \(i \in K\) in parallel}
                \State \(\mathbf{w_i(t)} \leftarrow \text{LocalUpdate}(i, \mathbf{w}(t-1))\)
            \EndFor
            \State Calculate \(\psi_i\) (equations \eqref{eq:entropy}, \eqref{eq:weight}).
            \State \(\mathbf{w}(t)=\sum^{K}_{i=1}\psi_i\mathbf{w}_i(t)\).
        \EndFor
        
        \State \textbf{Local execute}:
        \State \textbf{Function} LocalUpdate(\(i, \mathbf{w}\)):
        \For{\(e = 1, 2, \ldots, E\)}
            \For{\(b = 1, 2, \ldots, \lceil \frac{D_i}{B} \rceil\)}
                \State \(\mathbf{w} \leftarrow \mathbf{w} - \eta \nabla F(\mathbf{w})\)
            \EndFor
        \EndFor
        
        \State \textbf{return }\(\mathbf{w}\)
    \end{algorithmic}
\end{algorithm}

Algorithm \ref{alg:fedimp} outlines the FedImp algorithm, which iterates over \(T\) communication rounds. In each round, a randomly selected subset of (K) nodes participates. Key algorithm parameters include the local minibatch size \(B\), the number of local epochs \(E\), the learning rate \(\eta\), and the control parameter \(\tau\). A detailed step-by-step description follows.

\begin{itemize}
    \item Line 2: Initialize the global model parameters \(\mathbf{w(0)}\)
     \item Lines 3-10: Present main loop for communication rounds \(t\) from 1 to \(T\), 
     \ item Line 4: Select a random set of \(K\) nodes for the current round 
     \item Lines 5, 6: Loop through each node \(i\) in parallel and update its local model using the LocalUpdate function, which is further described later in the algorithm
     \item Line 8: Calculate \(\psi_i\) using equations (\ref{eq:entropy}) and (\ref{eq:weight})
     \item Line 9: Aggregate local models based on the calculated weights \(\psi_i\) to update the global model \(\mathbf{w}(t)\).
     \item Lines 12-18: Present the definition of the LocalUpdate function, which runs on each node \(i\).
     \item Line 13-17: Loop for local epochs \(e\)
     \item Lines 14-16: Loop for local mini-batches \(b\), updating the local model parameters \(\mathbf{w}\) using stochastic gradient descent with step size \(\eta\)
     \item  Line 18: Return the updated local model parameters. 
\end{itemize}

FedImp, like FedAvg and FedProx, aggregates local model parameters \( \mathbf{w}_i(t) \) instead of raw gradients. This choice supports multiple local epochs between rounds, letting models adapt more to local data before syncing. It avoids coordinating learning rates or batch stats across clients and ensures stable convergence in non-IID settings. For FedImp, parameter aggregation also aligns well with entropy-based weighting, without requiring extra gradient tracking.

Derived from equation \eqref{eq:weight}, when all participating nodes possess an identical quantity of data samples, the FedImp algorithm will allocate weights exclusively predicated on their respective impurity levels. Conversely, suppose there are variations in the sizes of data samples across participating nodes. In that case, FedImp will assign weights by considering both the impurity and the size of the data at each node.

\subsection{Complexity and  Convergence Analysis}

%To assess the scalability of FedImp, we analyze its computational complexity per client and communication overhead per round, comparing it with FedAvg and FedProx.

\subsubsection{Computational Cost Per Client}

Each client performs local training using mini-batch stochastic gradient descent (SGD) before sending the updated model to the server. The computational complexity per client primarily depends on the number of local epochs $E$, the number of samples per client $D_i$, and the model size (total number of parameters) $P$. 

For FedAvg and FedProx, the computational cost per client per communication round can be expressed as:
\begin{equation}
    \mathcal{O}(E D_i P).
\end{equation}

In FedImp, there is an additional step for computing the entropy-based impurity weight $\psi_i$ (equation (\ref{eq:weight})), which is derived from the entropy calculation for $D_i$ equation (\ref{eq:entropy})). Since the entropy computation requires iterating over all local samples to compute class proportions, its complexity is \( \mathcal{O}(D_i) \). Thus, the total complexity per client in FedImp is:
\begin{equation}
\mathcal{O}(ED_iP + D_i)
\end{equation}

In typical deep learning scenarios where the model size \( P \) is significantly larger than the number of classes \( C \), the \( \mathcal{O}(D_i) \) term becomes negligible compared to \( \mathcal{O}(ED_iP) \). As such, the overall computational complexity of FedImp remains $\mathcal{O}(ED_iP)$.

In practical implementations, the impurity computation per client involves a single pass over local labels to count class frequencies, followed by a computation of entropy using these counts. This can be implemented efficiently using a histogram-based class counter, requiring negligible memory and minimal additional computation. Even on resource-constrained devices such as IoT sensors or mobile phones, this overhead is expected to be minimal, especially when compared to the cost of local model training (i.e., forward and backward passes in deep networks). Thus, FedImp remains feasible for deployment in low-resource federated environments.

\subsubsection{Communication Overhead Per Round}
The communication overhead in FL is primarily determined by the number of parameters transmitted between clients and the central server. In each round, every client sends its model update $\mathbf{w}_i$ (a vector of size $P$) to the server, which aggregates updates and transmits the global model $\mathbf{w}$ back to the clients.

For FedAvg and FedProx, each client transmits and receives $\mathbf{w}$ of size $P$, leading to a communication cost per client per round of $\mathcal{O}(P)$.

For FedImp, an additional scalar value $\psi_i$ must be transmitted for weighting. Since $\psi_i$ is a single floating-point value, its communication overhead is negligible compared to the model parameters. Thus, the total communication cost for FedImp remains $\mathcal{O}(P)$.

In summary, FedImp introduces an additional entropy calculation step, but this does not change its overall computational complexity compared to FedAvg and FedProx. Additionally, the communication overhead remains unchanged since the weighting factor $\psi_i$ is a negligible scalar value.

\subsubsection{Convergence of FedImp}
The convergence of FedImp is driven by its impurity-based weighting mechanism, which ensures that clients with more diverse and representative data contribute more significantly to the global model. Unlike FedAvg, which weights updates solely based on dataset size, FedImp prioritizes clients with higher entropy (diverse class distributions), promoting balanced and informative aggregation.

In particular, FedImp tends to exhibit improved convergence behavior under the following conditions:
\begin{itemize}
    \item The number of classes $C$ is moderate relative to dataset size, allowing entropy to effectively reflect class diversity.
    \item There exists a mix of balanced and imbalanced clients, where entropy weighting can help differentiate and amplify the impact of balanced clients.
    \item The local training process follows standard SGD or its variants with bounded gradients and smooth loss functions.
\end{itemize}

These conditions align with typical non-IID federated settings where some nodes possess skewed or low-diversity data. By emphasizing updates from more representative clients, FedImp reduces bias and accelerates convergence toward a global optimum. While this study focuses on empirical validation of convergence improvements, a formal theoretical convergence bound for FedImp remains an important direction for future work.

\section{Experiments and Results}
This section evaluates our proposed FedImp algorithm and compares its performance with FedAvg, FedProx, and FedAdp. 
\label{sec:experiment}

\subsection{Dataset, Federated Learning Models, and Testing Scenario }
\subsubsection{Creation of Non-IID  Data}

The experiments are carried out on two datasets:
\begin{itemize}
\item EMNIST \cite{deng2012mnist}: A 28x28 pixel handwritten character dataset, we use the Balanced split (47 characters, equal samples per class) with 112,800 training and 18,800 testing samples.
\item CIFAR-10 \cite{Krizhevsky09learningmultiple}: 60,000 32x32 color images across 10 classes (6,000 per class), with 50,000 training and 10,000 testing images.
\end{itemize}

These datasets provide a standardized, well-understood benchmark, allowing direct comparisons with existing FL methods while ensuring the experiments remain computationally efficient. Future work can extend FedImp’s evaluation to larger and more complex datasets.

To simulate a diverse population across client nodes, the same approach as in \cite{reddi2020adaptive} is used with some modifications. Each client has an associated multinomial distribution over classes, which is drawn from a symmetric Dirichlet distribution, \(q \sim \text{Dir}(\theta)\). The positive number \(\theta > 0\) serves as a concentration parameter controlling the level of balance among classes. A sufficiently large \(\theta\) contributes to the creation of a balanced dataset, whereas a sufficiently small \(\theta\) results in an imbalanced dataset. When \(\theta\) increases to infinity, all clients exhibit identical distributions. Conversely, as \(\theta\) approaches 0, each client exclusively possesses examples from a single class chosen at random.

\begin{algorithm}[ht!]
    \caption{Data partition}
    \label{alg:data_partition}
    \begin{algorithmic}[1]
        \Require \(N_b\), \(N_{imb}\), \(\theta_{b}\), \(\theta_{imb}\), \(M\)
        
        \For{\(i = 1, 2, \ldots, N_b+N_{imb}\)}
            \If{\(i \le N_b\)}
                \State Sample \(q\sim \text{Dir}(\theta_b, C)\)
            \Else
                \State Sample \(q\sim \text{Dir}(\theta_{imb}, C)\)
            \EndIf
            
            \State \(D_i=\text{\O}\)
            
            \For{\(j = 1, 2, \ldots, M\)}
                \State Sample \(y \in C \text{ with probability } q\)
                \State Sample randomly \(x \in S_y\)
                \State \(D_i = D_i \cup {(x,y)}\)
                \State \(S_y = S_y \setminus {(x,y)}\)
                
                \If{\(|S_y|=0\)}
                    \State \(C \setminus y\)
                    \State \(q \leftarrow \text{ReNormalize}(q,y)\)
                \EndIf
            \EndFor
        \EndFor

        \State \textbf{Support function}:
        \State \textbf{Function} ReNormalize(\(q=(p_1,p_2,...,p_C),y\)):
            \State \quad \(p_y=0\)
            \State \quad \(a=\sum_{i=1}^C p_i\)
            \State \quad \(q=q/a\)
            \State \quad \Return \(q\)
        
    \end{algorithmic}
\end{algorithm}

Algorithm \ref{alg:data_partition} outlines the process of partitioning the training set with a specified number of balanced \(N_b\) and imbalanced \(N_{imb}\) data nodes. Each node has \(M\) samples. The distribution of data involves sampling class probabilities from a Dirichlet distribution with concentration parameters \(\theta_b\) and \(\theta_{imb}\) for balanced and imbalanced nodes, respectively. The procedures of Algorithm \ref{alg:data_partition} are summarized below.
\begin{itemize}
    \item Lines 1 to 18: Loop over each node \(i\) from 1 to (\(N_b+N{imb}\)) to generate its data sampling.
    \item Lines 2 to 5: If \(i\) is less than or equal to \(N_b\), the associated multinomial distribution \(q\) is obtained from a Dirichlet distribution with concentration parameter \(\theta_1\) and \(C\) categories. Otherwise, otherwise, it is sampled with concentration parameter \(\theta_2\).
    \item Lines 8 to 17: For each node \(i\), loop through \(M\) samples with the following steps:
    \item Line 9 to  12: Sample a class label \(y\) based on the probability distribution \(q\), randomly select a sample \((x, y)\) from the remaining samples of class \(y\) in the dataset \(S\), add the sample \((x, y)\) to the dataset \(D_i\), remove the selected sample from the dataset \(S_y\) of class \(y\).
    \item Lines 13 to 16: If there are no more samples for class \(y\), remove \(y\) from the set of classes \(C\), and re-normalize the class probabilities \(q\). The  \(ReNormalize(q,y)\) function ensures that the class probabilities \(q\) are normalized after removing a class. This partitioning method guarantees all samples from the original dataset are sampled and prevents any duplication of samples within nodes.
\end{itemize}

For the experiments, we set \(\theta_b=100\) and \(\theta_imb=0.01\). The entire dataset is used, resulting in \(M = \frac{S}{N_b+N{imb}}\) samples allocated to each node. To evaluate the performance of the global model, we follow standard federated learning evaluation practice by using a centralized test set. Specifically, we use the entire original test set of each dataset (EMNIST and CIFAR-10) as a global benchmark, and after each communication round, the updated global model is evaluated on this fixed test set. This ensures consistency and comparability across different algorithms and data distribution scenarios.

\begin{figure*}[t]
  \centering
  \subfloat[MLP for EMNIST]{\includegraphics[scale=0.07]{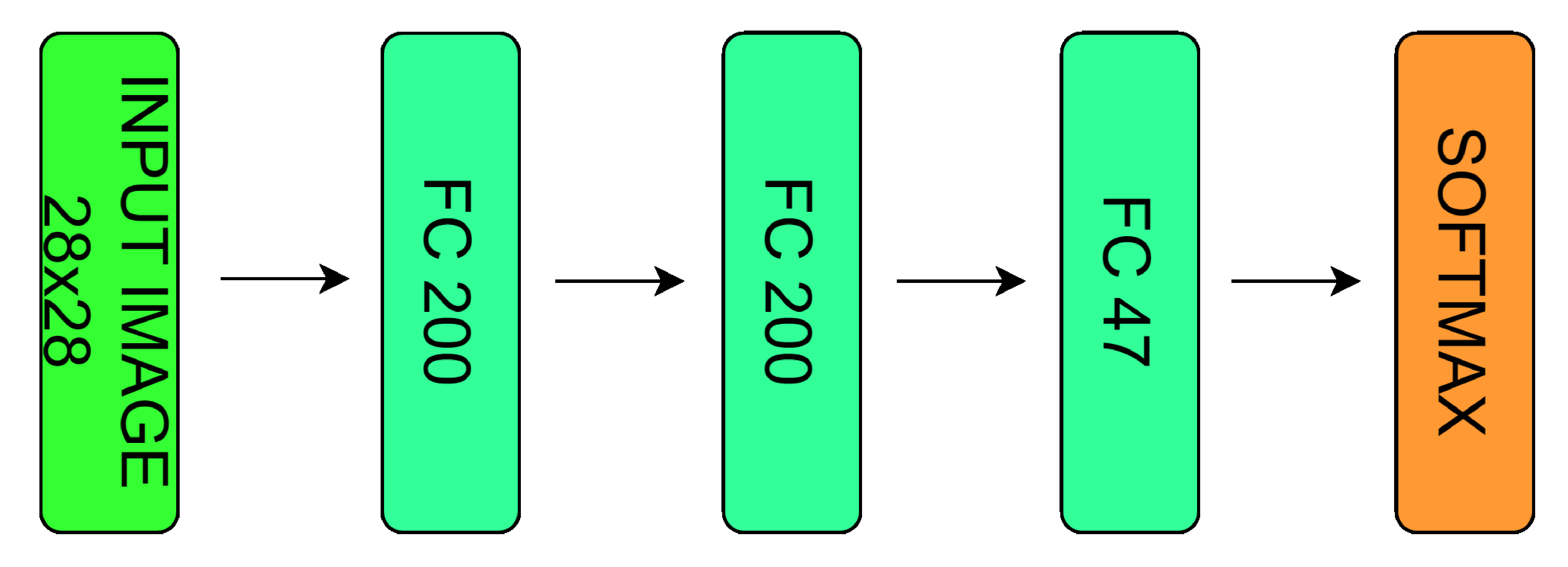}}
  \hfill
  \subfloat[2-Layer CNN for EMNIST]{\includegraphics[scale=0.07]{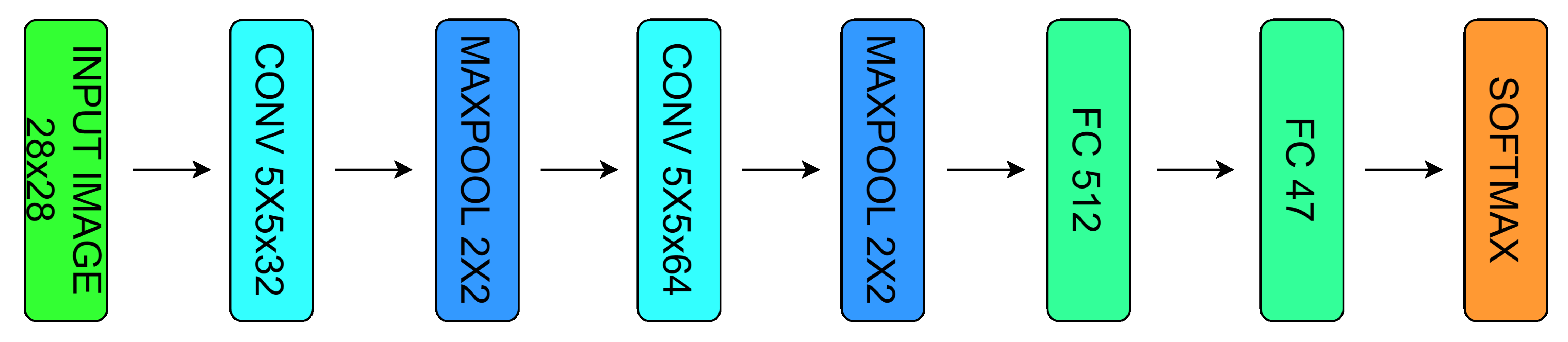}}
  \\
  \subfloat[2-Layer CNN for CIFAR10]{\includegraphics[scale=0.07]{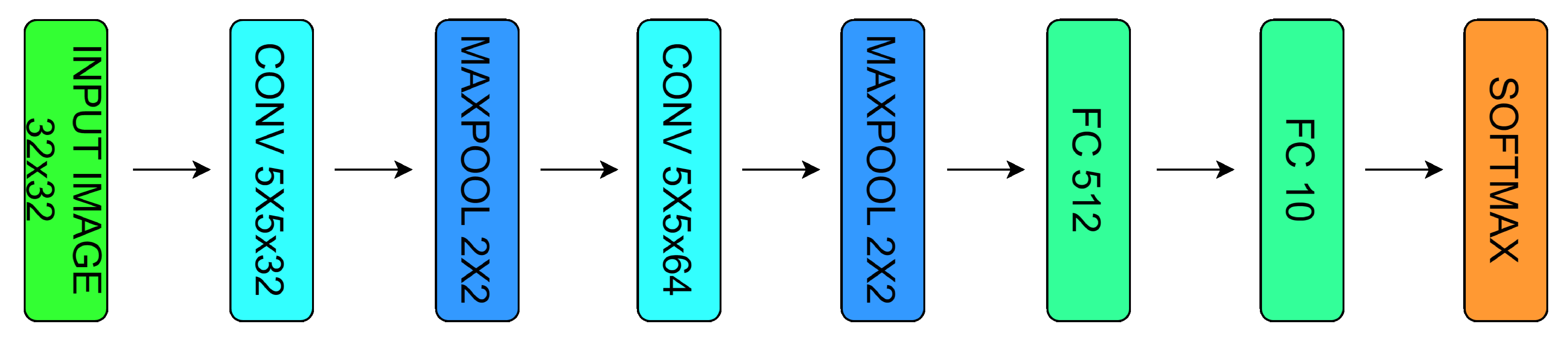}}
  \hfill
  \subfloat[4-Layer CNN for CIFAR10]{\includegraphics[scale=0.07]{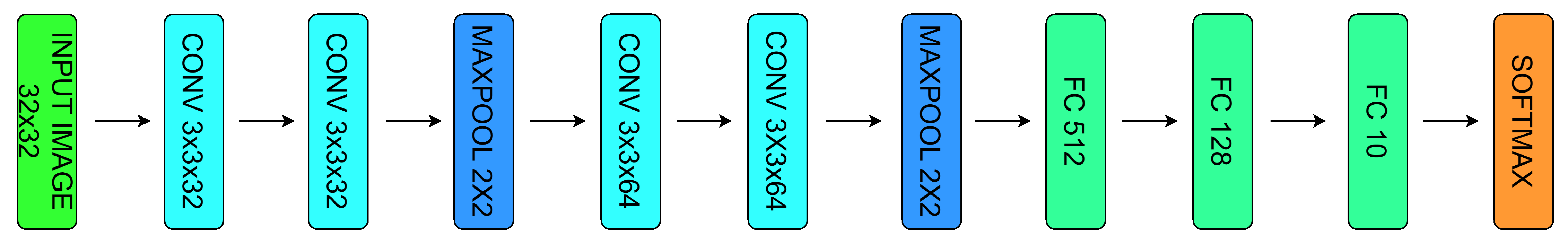}}
  \caption{Models architectures: (a) MLP and (b) 2-layer CNN for EMNIST; (c) 2-layer CNN and (d) 4-layer CNN for CIFAR10}
  \label{fig:model_architecture}
\end{figure*}

\subsubsection{Structures of the Federated Learning Models}
The experiments involve the FL training of four artificial neural networks (ANN) models. The first two are an MLP model and a 2-layer CNN model for EMNIST data. For the CIFAR10 data, we test 2-layer CNN and 4-layer CNN models. Their architectures (see Figure \ref{fig:model_architecture}) are as follows: 
\begin{itemize}
    \item MLP for EMNIST: This model has two fully connected hidden layers of 200 units using ReLU activation. The input layer is set to fit the EMNIST image size of 28x28, while the output layer contains 47 neurons with Softmax activation for EMNIST characters classification.
    
    \item 2-layer CNN for EMNIST: This model has two 5x5 convolutional layers (the first with 32 channels, the second with 64, each followed with 2x2 max pooling) and a fully connected layer with 512 units using ReLu activation. The input and output layers have the same format as in the MLP model for the EMNIST data.
 
     \item 2-layer CNN for CIFAR10: The main components of this model are the same as in the 2-layer CNN for EMNIST, except that the input layer has more neurons to match the  CIFAR10 input image size. The output layer of this model is changed to contain 10 neurons for classifying CIFAR10 images.
     
    \item 4-layer CNN for CIFAR10: This model has four convolutional layers (32, 32, 64, and 64 channels, 3x3 kernels, ReLU activation), with max pooling after the second and fourth layers.  Two fully connected layers (512 and 128 units, 40\% dropout, ReLU) follow. Input and output layers match the 2-layer CNN model for CIFAR10.
    %This model contains four 3x3 convolutional layers with 32, 32, 64, and 64 channels, respectively) and ReLu activation. The second and the last convolutional layers are accompanied by 2x2 max pooling. After the CNN layers, the model uses two subsequent fully connected layers with 512 and 128 units, each including dropout layers with a 40\% dropout rate and ReLu activation. The model has the same structures of the input and output layer as in the 2-layer CNN model for the CIFAR10 dataset.
\end{itemize}
%All three models have an output layer of 10 units, which uses Softmax activation  %The EMNIST dataset will be used to train FL models of the MLP and 2-layer CNN networks while the CIFAR-10 will be used for training models of 2-layer and 4-layer CNN networks.
%The input layer of each network will be designed and changed accordingly to match the input data. 

\subsubsection{Testing scenarios}

To compare the FedImp algorithm with FedAvg, FedProx, and FedAdp algorithms, the testing scenarios are devised with changes in the number of nodes with balanced data and nodes with imbalanced data. For each of the two datasets, EMNIST and CIFAR10, four testing scenarios are generated for evaluation as follows:
\begin{itemize}
    \item 7 balanced nodes + 3 imbalanced nodes (non-IID) 
    \item 5 balanced nodes + 5 imbalanced nodes (non-IID)
    \item 3 balanced nodes + 7 imbalanced nodes (non-IID)
    \item 3 balanced nodes + 7 identically distributed and imbalanced nodes (extreme skew non-IID)
\end{itemize}

The data in the first three non-IID scenarios is partitioned using Algorithm \ref{alg:data_partition} with \(\theta_b=100\), \(\theta_{imb}=0.01\).

The last scenario represents the case of FL with 3 balanced nodes + 7 identically distributed and imbalanced nodes. In this case, the training data of the imbalanced nodes are randomly selected from just a few classes of data instead of from all classes as in the other case. This is an extreme non-IID scenario where the data is highly skewed to just a few classes. In each of the 3 balanced nodes, \(M\) samples will be randomly sampled from the original training set, while in the 7 identically distributed and imbalanced nodes, \(M\) samples will be randomly sampled from a subset of only 2 classes for CIFAR and only 10 digits for EMNIST (omitting letters). Specifically, for the EMNIST dataset, the 10 digits (classes 0–9) were used as the fixed subset across all runs to ensure consistency. For the CIFAR dataset, a random subset of 2 classes was chosen for each run. In this scenario, the sampling process adheres to a mechanism of sampling with replacement, which ensures a sufficient number of samples per node.

\subsubsection{Pre-processing and Hyperparameter Configuration}

\textit{Input Image Pre-processing}: Several image transformations are executed before training the  FL models. For the EMNIST images, the normalization is applied to the pixel arrays to have a mean of 0.5 and a standard deviation of 0.5. The CIFAR-10 images are processed with random cropping of size 32 with padding of 4 pixels, random horizontal flipping, and normalization of the pixels arrays to have a mean of 0.5 and a standard deviation of 0.5.

\textit{Hyperparameter}: The experiments employ all available nodes for both training at each round (i.e., a utilization fraction of 1.0). During each round, client nodes train their local models individually for one epoch with a batch size of 100. The local model training employs the Stochastic Gradient Descent (SGD) optimization algorithm to minimize cross-entropy loss. The learning rate is set to 0.1, and a decay rate of 0.995 is applied after each communication round. Additionally, the weight \(\mu\) of the proximal term in FedProx is set to 0.1, the constant \(\alpha\) in the non-linear mapping function of FedAdp is set to 5, the temperature parameter \(\tau\) in the weight forming function of FedImp is set to 0.7. After each round, the global model's performance is assessed using the test sets from all participating nodes. The values for key hyperparameters such as \(\mu = 0.1\) for FedProx, \(\alpha = 5\) for FedAdp, and \(\tau = 0.7\) for FedImp were selected based on preliminary experiments and empirical tuning on a small validation split derived from the training data. The goal was to ensure that each method operated under reasonably optimized settings. For FedProx and FedAdp, we also consulted values recommended in their original papers. We fixed these values across all scenarios to ensure fairness and reproducibility in our comparisons.

\subsection{Results and Discussion} 
To compare the convergence rates of FedImp, FedAvg, FedProx, and FedAdp algorithms, we measure the number of communication rounds needed to reach convergence.

\begin{figure}
  \centering
  \subfloat{\includegraphics[clip,trim={0 0 0 0},width=0.45\linewidth]{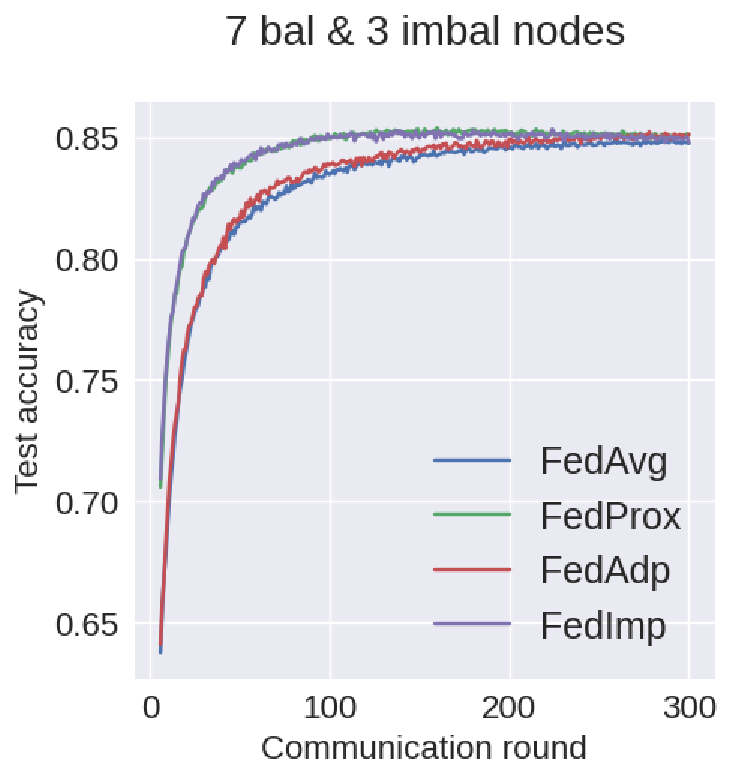}}
  \hfill
  \subfloat{\includegraphics[width=0.5\linewidth]{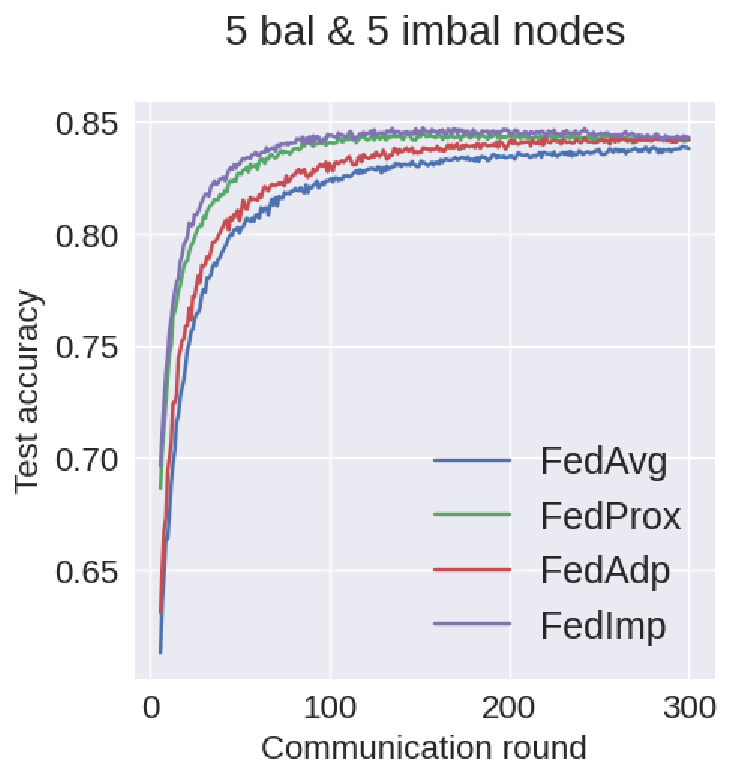}}
  \hfill
  \subfloat{\includegraphics[width=0.5\linewidth]{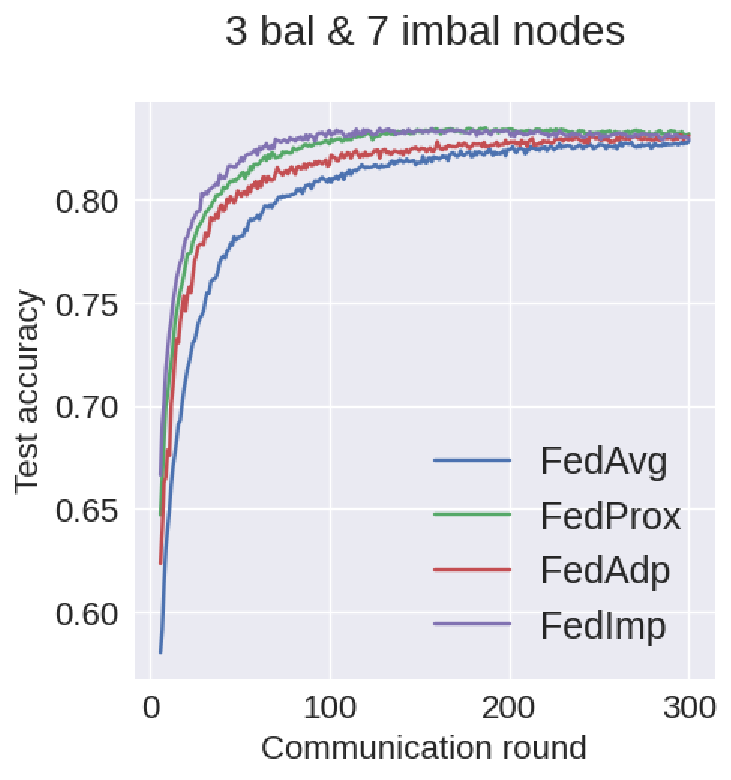}}
  \hfill
  \subfloat{\includegraphics[width=0.5\linewidth]{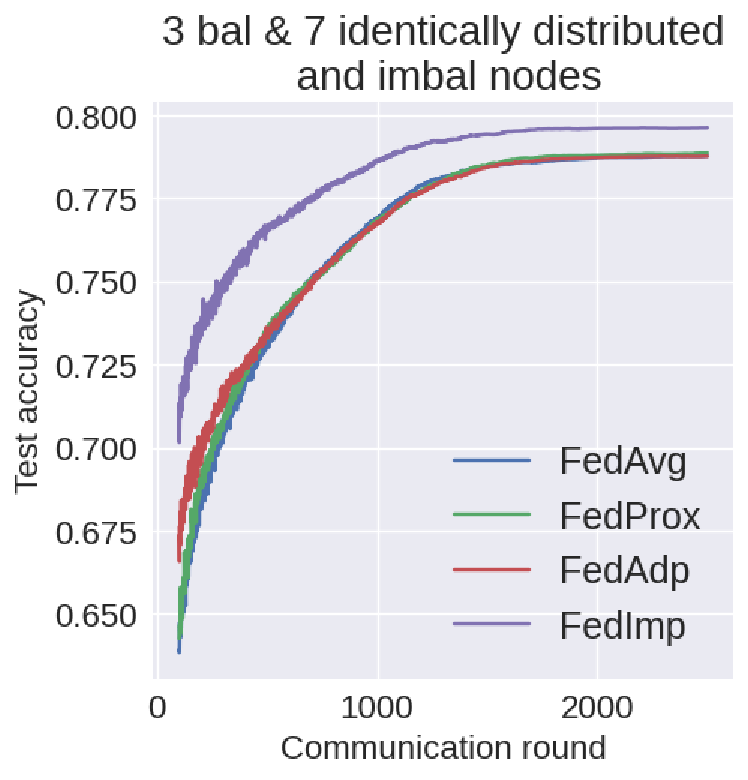}}
  \caption{\textbf{MLP} model for \textbf{EMNIST} results of test accuracy over communication rounds on FedAvg, FedProx, FedAdp, and FedImp algorithms with different levels of heterogeneous data distribution over participating nodes.}
  \label{fig:emnist_1}
\end{figure}

\begin{table}
\centering
\caption{The number of communication rounds for each FL algorithm to reach over target test accuracy with the \textbf{MLP} model for \textbf{EMNIST} data. N/A indicates that the algorithm cannot achieve the target accuracy.}
\label{tab:emnist_1}
\begin{tabular}{|m{2cm}|c|c|c|c|c|} \hline
\textbf{Testing scenario} & ACC & \textbf{FedAvg} & \textbf{FedProx} & \textbf{FedAdp} & \textbf{FedImp} \\ \hline
7 bal + 3 imbal & 83\% & 79 & 33 & 71 & 35 \\ %\hline
5 bal + 5 imbal & 83\% & 132 & 56 & 94 & 47 \\ %\hline
3 bal + 7 imbal & 83\% & N/A & 108 & 234 & 78 \\ %\hline
3 bal + 7 iden. dist. imbal & 79\% & N/A & N/A & N/A & 1126 \\ \hline
\end{tabular}
\end{table}

\subsubsection{Results on EMNIST data}
\textit{MLP model for EMNIST results:} Figure \ref{fig:emnist_1} illustrates the performance of FedAvg, FedProx, FedAdp, and FedImp algorithms in terms of test accuracy across communication rounds under various heterogeneous data distribution scenarios among participating nodes, employing the MLP model. Table \ref{tab:emnist_1} summarizes the number of communication rounds required to reach a certain accuracy level, chosen based on the maximum accuracy of all four algorithms on the EMNIST dataset in each data distribution scenario. Across all scenarios, the proposed FedImp consistently displays a fast convergence rate compared to other algorithms. The observed results are as follows:
\begin{itemize}
    \item In the scenario featuring 7 balanced nodes and 3 imbalanced nodes, FedImp closely aligns with FedProx; they take 35 and 33 rounds, respectively, to surpass 83\% accuracy. Both FedImp and FedProx outperform FedAvg (79 rounds) and FedAdp (71 rounds).
    \item In the FL with 5 balanced nodes and 5 imbalanced nodes, FedImp demonstrates a notable advancement, which takes only 47 communication rounds to achieve 83\% or higher accuracy. FedProx, FedAdp, and FedAvg require 56, 94, and 132 rounds, respectively. In this case, FedImp has shown large reductions in communication rounds of approximately 16.1\%, 50\%, and 64.4\% less than FedProx, FedAdp, and FedAvg.
    \item In the scenario of 3 balanced nodes and 7 imbalanced nodes, FedProx and FedAdp require 108 and 234 rounds to reach the required 83\% accuracy. FedAvg cannot reach target accuracy in this case. It can approach nearly 83\% after more than 300 rounds of communication. The proposed FedImp algorithm exhibits clear superiority over all other algorithms, as it takes just 78 rounds to reach the target accuracy. FedImp has required fewer rounds by approximately 27.8\%, 66.7\% than FedProx and FedAdp.
    \item In the last case of 3 balanced nodes + 7 identically distributed and imbalanced nodes, FedImp has clearly outperformed all other algorithms, as it is the only one that can converge at 79\% accuracy. All three other algorithms (FedProx, FedAdp, and FedAvg) only obtained an accuracy of around 77\%.
\end{itemize}

\begin{figure}
  \centering
  \subfloat[]{\includegraphics[width=0.5\linewidth]{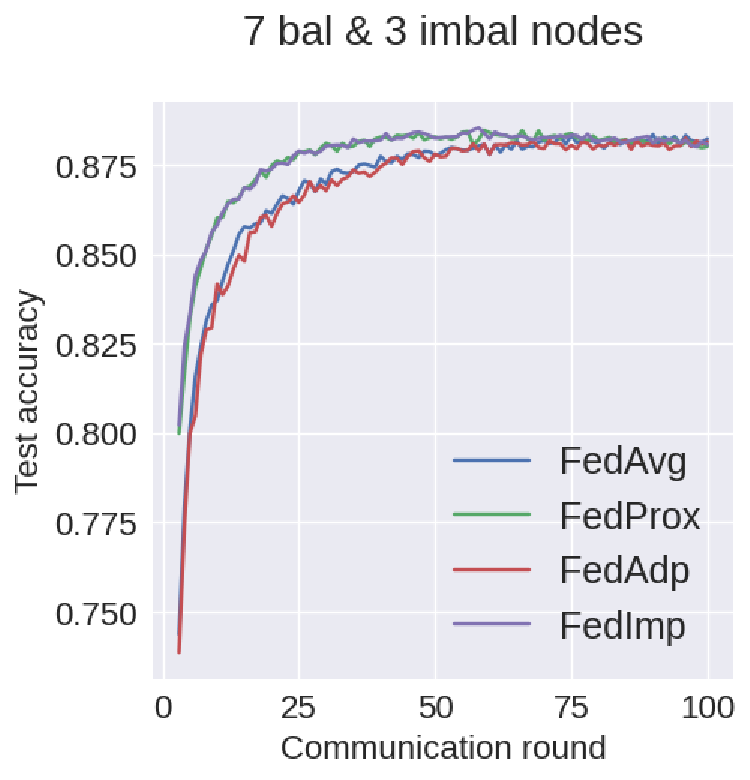}}
  \hfill
  \subfloat[]{\includegraphics[width=0.5\linewidth]{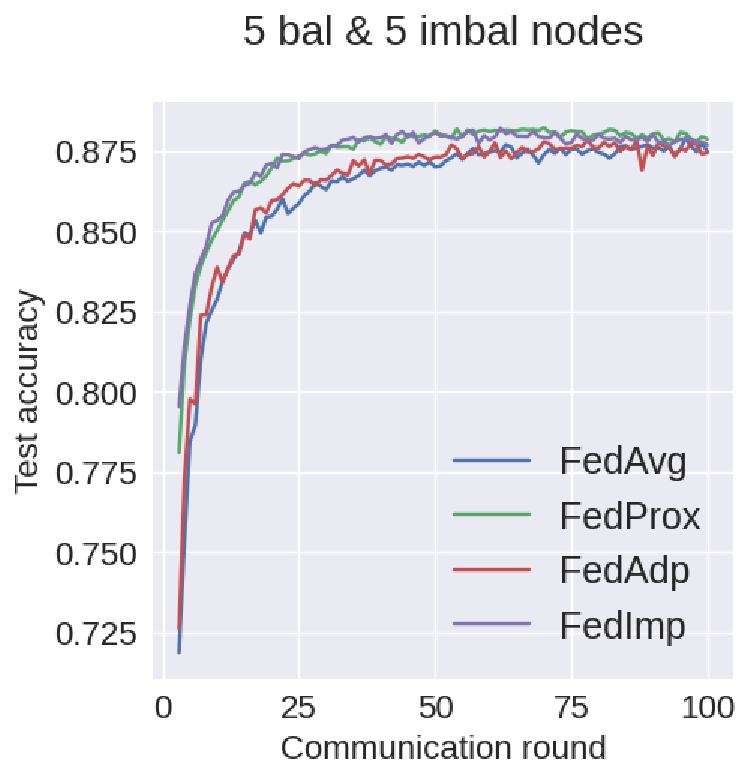}}
  \hfill
  \subfloat[]{\includegraphics[width=0.5\linewidth]{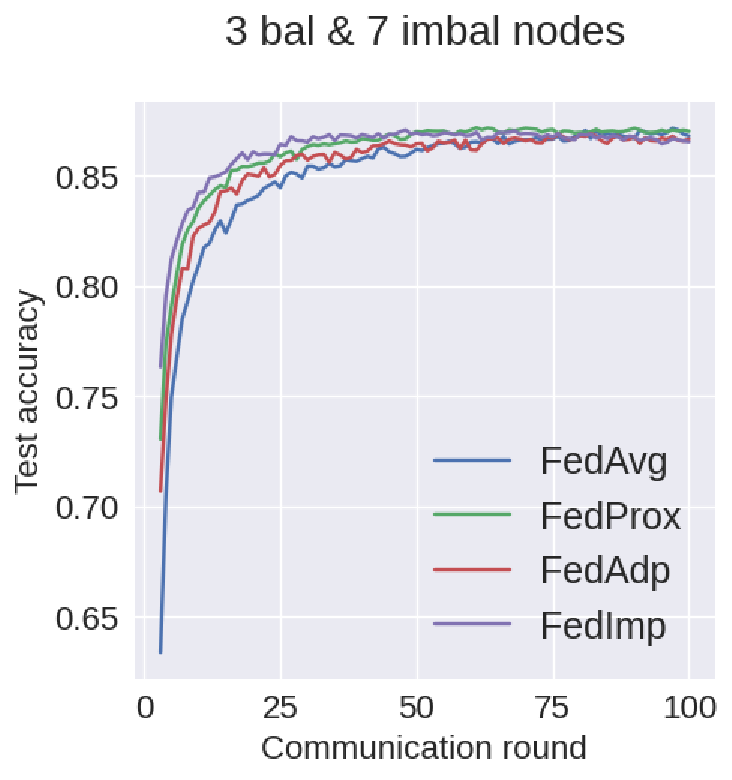}}
  \hfill
  \subfloat[]{\includegraphics[width=0.5\linewidth]{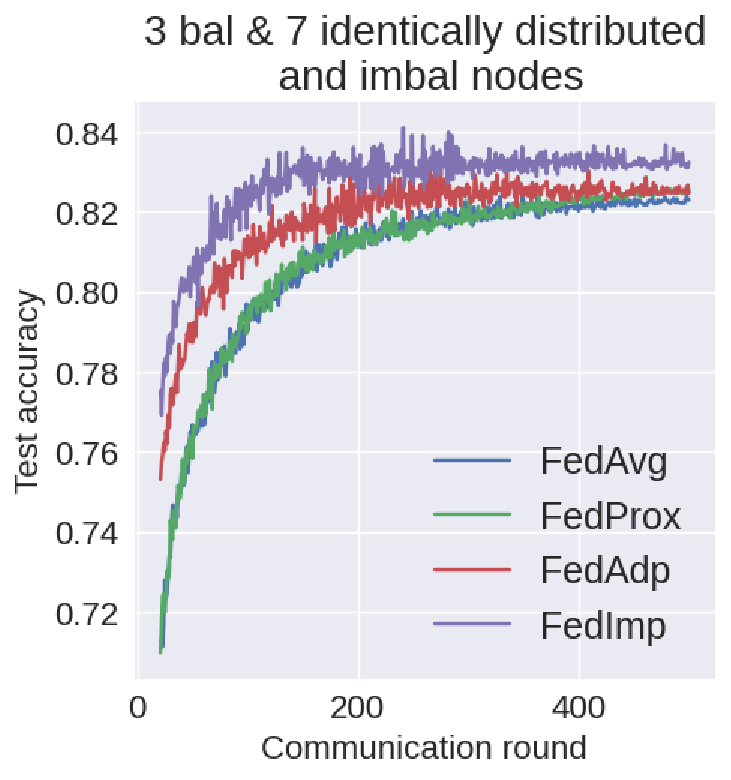}}
  \caption{\textbf{2-layer CNN} model for \textbf{EMNIST} results on test accuracy over communication rounds of FedAvg, FedProx, FedAdp, and FedImp with different levels of heterogeneous data distribution over participating nodes.}
  \label{fig:emnist_2}
\end{figure}

\begin{table}
\centering
\caption{The number of communication rounds for FL algorithms to reach over target test accuracy with the \textbf{2-layer CNN} model for \textbf{EMNIST} data. N/A indicates that the algorithm cannot achieve the target accuracy.}
\label{tab:emnist_2}
\begin{tabular}{|m{2cm}|c|c|c|c|c|} \hline
\textbf{Testing scenario} & ACC & \textbf{FedAvg} & \textbf{FedProx} & \textbf{FedAdp} & \textbf{FedImp} \\ \hline
7 bal + 3 imbal & 87\% & 26 & 17 & 27 & 18 \\ %\hline
5 bal + 5 imbal & 87\% & 41 & 21 & 35 & 19 \\ %\hline
3 bal + 7 imbals & 87\% & 81 & 50 & N/A & 48 \\ %\hline
3 bal + 7 iden. dist. imbal & 83\% & N/A & N/A & 201 & 115 \\ \hline
\end{tabular}
\end{table}

\textit{2-Layer CNN for EMNIST results:}
Figure \ref{fig:emnist_2} illustrates test accuracy over communication rounds for all four algorithms, considering diverse scenarios of heterogeneous data distribution among participating nodes and utilizing the 2-layer CNN model. Table \ref{tab:emnist_2} presents the communication rounds required for each algorithm to achieve an accuracy level determined by the best accuracy of all algorithms for the data distribution scenario. The results have shown the advancement of FedImp over other algorithms.
\begin{itemize}
    \item In the scenario of 7 balanced nodes and 3 imbalanced nodes, FedImp and FedProx reach the target accuracy (87\%) after 18 and 17 rounds, respectively, whereas FedAvg and FedAdp require 26 and 29 rounds. 
    \item In the scenario of 5 balanced nodes and 5 imbalanced nodes, FedImp achieves the target accuracy (87\%) in 19 rounds, indicating a slight improvement over FedProx (needs 21 rounds), and demonstrating superiority over FedAvg (needs 41 rounds), and FedAdp (needs 35 rounds), with reductions of 53.7\% and 45.7\% in that order. 
    \item The results in the case of FL with 3 balanced nodes and 7 imbalanced nodes further highlight the efficiency of FedImp. It achieves convergence in 48 rounds and demonstrates a remarkable reduction in the number of rounds, with an efficiency gain of 4\% over FedProx and an even more substantial improvement of about 40.7\% over FedAvg. In this scenario, FedAdp fails to reach the 87\% accuracy target.
    \item For the case of 3 balanced nodes + 7 identically distributed and imbalanced nodes, the target accuracy is 83\%. Only FedImp (115 rounds) and FedAdp (201 rounds) can achieve this accuracy level. However, FedAdp later converged to just around 82\% accuracy, while FedImp still maintained its higher performance.
    
\end{itemize}

\begin{figure}[t]
  \centering
  \subfloat[]{\includegraphics[width=0.5\linewidth]{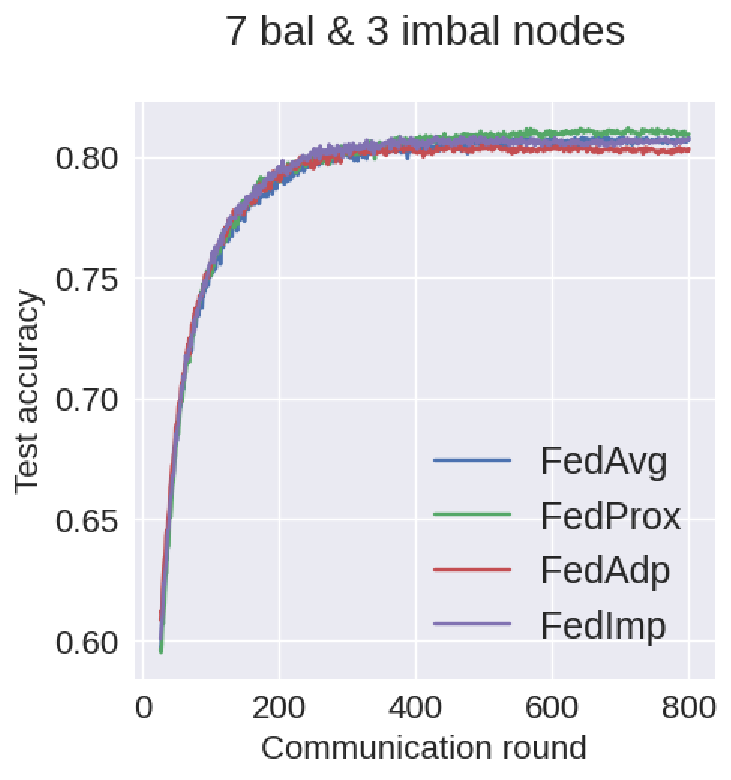}}
  \hfill
  \subfloat[]{\includegraphics[width=0.5\linewidth]{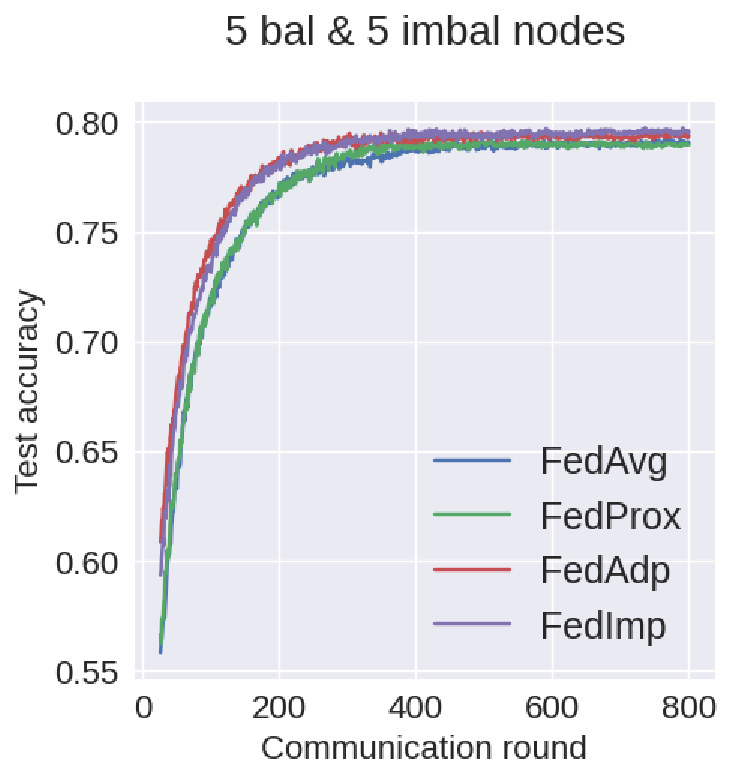}}
  \hfill
  \subfloat[]{\includegraphics[width=0.5\linewidth]{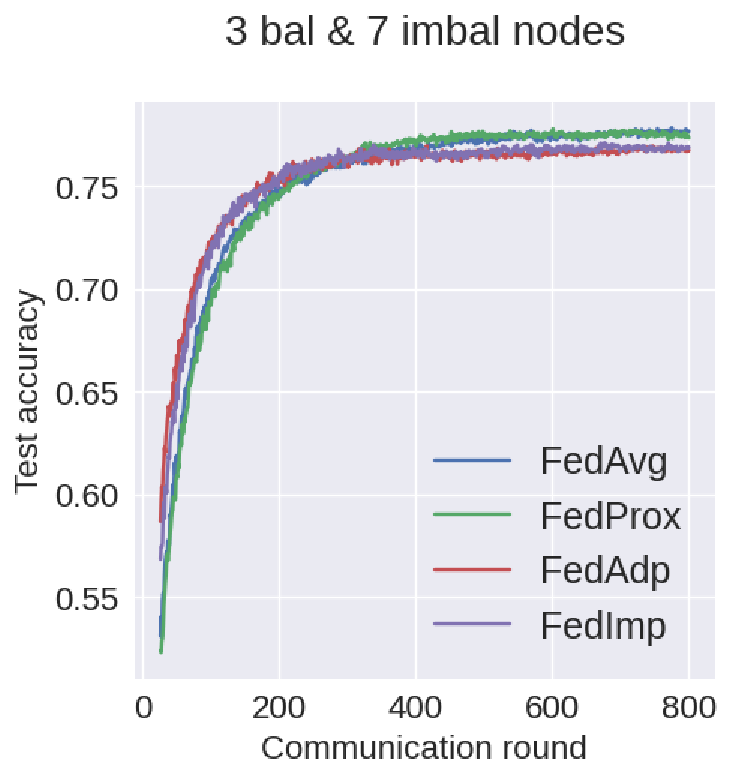}}
  \hfill
  \subfloat[]{\includegraphics[width=0.5\linewidth]{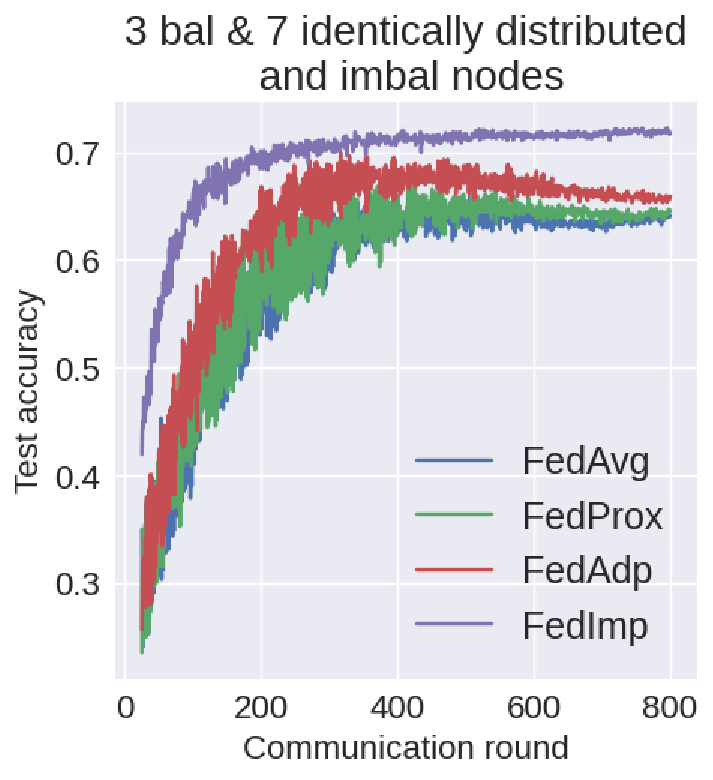}}
  \caption{\textbf{2-layer CNN} model for \textbf{CIFAR-10} results on test accuracy over communication rounds of FedAvg, FedProx, FedAdp, and FedImp with different levels of heterogeneous data distribution over participating nodes.}
  \label{fig:cifar_1}
  \vspace{-5pt}
\end{figure}

\begin{table}[t]
\centering
\caption{The number of communication rounds for FL algorithms to reach over target test accuracy with the \textbf{2-layer CNN} model for \textbf{CIFAR-10} data. N/A indicates that the algorithm cannot achieve the target accuracy.}
\label{tab:cifar_1}
\begin{tabular}{|m{2cm}|c|c|c|c|c|} \hline
\textbf{Testing scenario} &ACC & \textbf{FedAvg} & \textbf{FedProx} & \textbf{FedAdp} & \textbf{FedImp} \\ \hline
7 bal + 3 imbal & 80\% & 247 & 237 & 231 & 235 \\ %\hline
5 bal + 5 imbal & 79\% & 452 & 363 & 255 & 252 \\ %\hline
3 bal + 7 imbal & 76\% & 270 & 250 & 220 & 211 \\ %\hline
3 bal + 7 iden. dist. imbal & 72\% & N/A  & N/A & N/A & 408 \\ \hline
\end{tabular}
\end{table}

\subsubsection{Results on CIFAR-10 Data}
\textit{2-Layer CNN for CIFAR-10 results:}
Figure \ref{fig:cifar_1} demonstrates the evolving test accuracy across communication rounds for FedAvg, FedProx, FedAdp, and FedImp, taking into account diverse scenarios of heterogeneous data distribution among participating nodes and employing the 2-layer CNN model. 
%Notably, in the case with only 3 imbalanced data nodes, there is not a significant difference in the performance of the algorithms  In the two scenarios characterized by a higher number of imbalanced data nodes, both FedImp and FedAdp exhibit a superior convergence rate compared to FedAvg and FedProx. 
Table \ref{tab:cifar_1} presents the results for the number of communication rounds required by FedAdp, FedAvg, and FedImp algorithms to achieve the target accuracy. 
\begin{itemize}
    \item In the scenario with 7 balanced nodes and 3 imbalanced nodes, FedImp takes 235 rounds to surpass 80\%, while FedAvg and FedProx require a slightly higher number of rounds, namely 247 and 237 in the order given. Nonetheless, FedImp is slightly slower than FedAdp, which takes 231 rounds. 
    \item For the case with 5 balanced nodes and 5 imbalanced nodes, FedImp outperforms FedAvg, FedProx, and FedAdp by converging at 79\% in 252 rounds, while FedAvg, FedProx, and FedAdp require 452, 363, and 255 rounds, respectively. This demonstrates a reduction of rounds by about 44.2\%, 30.6\%, and 1.2\% for FedImp compared to FedAvg, FedProx, and FedAdp, respectively.
    \item  In the scenario of 3 balanced nodes and 7 imbalanced nodes, FedImp again showcases its efficiency by achieving convergence to 76\% accuracy in 211 rounds, while FedAvg, FedProx, and FedAdp require 270, 250, and 220 rounds, respectively. This reflects a reduction of rounds by about 21.9\%, 15.6\%, and 4.1\% when comparing FedImp to FedAvg, FedProx, and FedAdp in that order.
    \item In the scenario of 3 balanced nodes + 7 identically distributed and imbalanced nodes, FedImp is the only algorithm that can converge at 72\%. All three other algorithms can only obtain an accuracy of around 65\%.
\end{itemize}

\begin{figure}[t]
  \centering
  \subfloat[]{\includegraphics[width=0.5\linewidth]{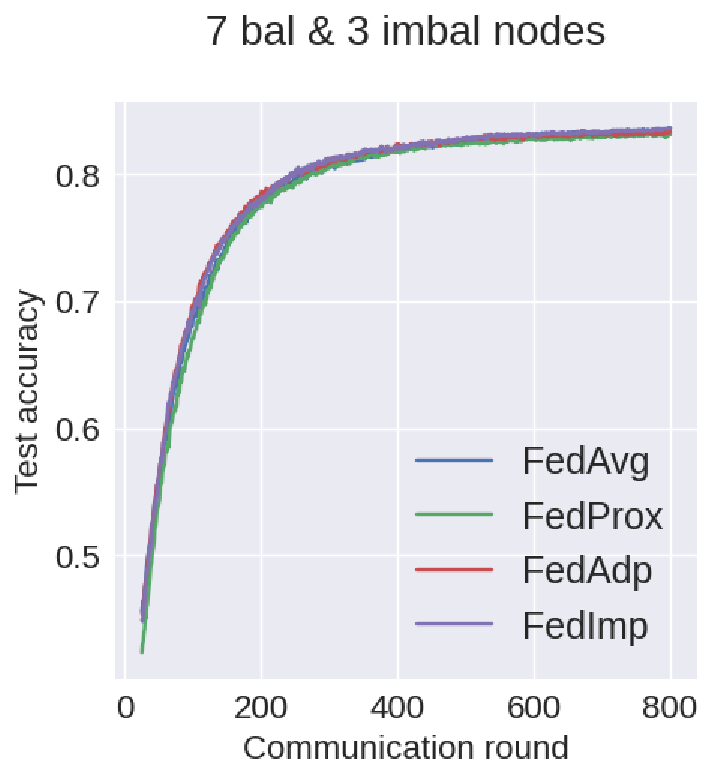}}
  \hfill
  \subfloat[]{\includegraphics[width=0.5\linewidth]{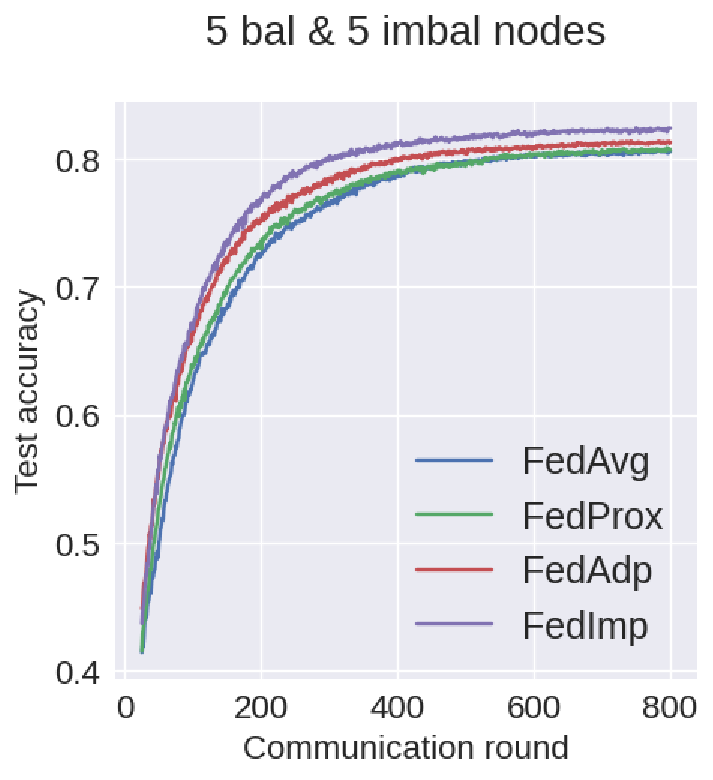}}
  \hfill
  \subfloat[]{\includegraphics[width=0.5\linewidth]{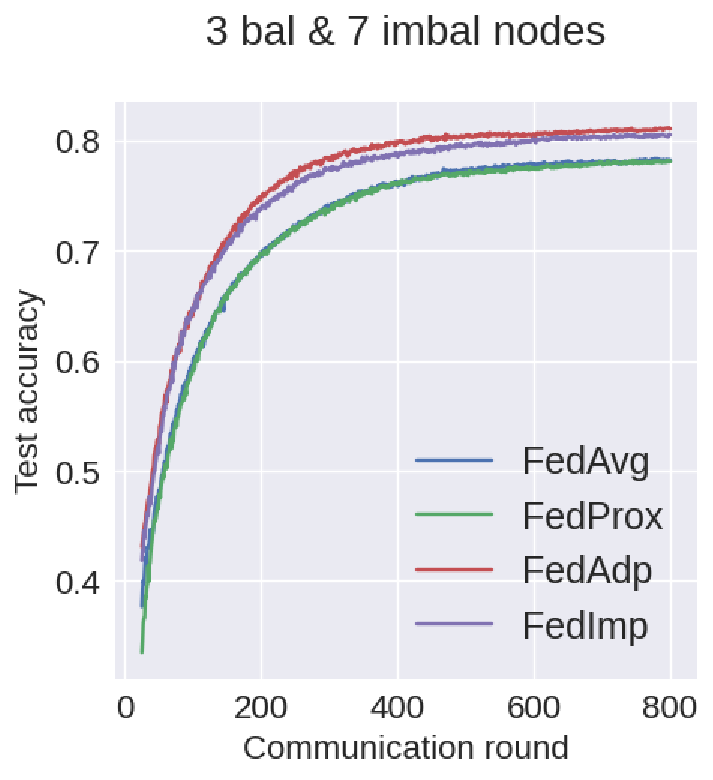}}
  \hfill
  \subfloat[]{\includegraphics[width=0.5\linewidth]{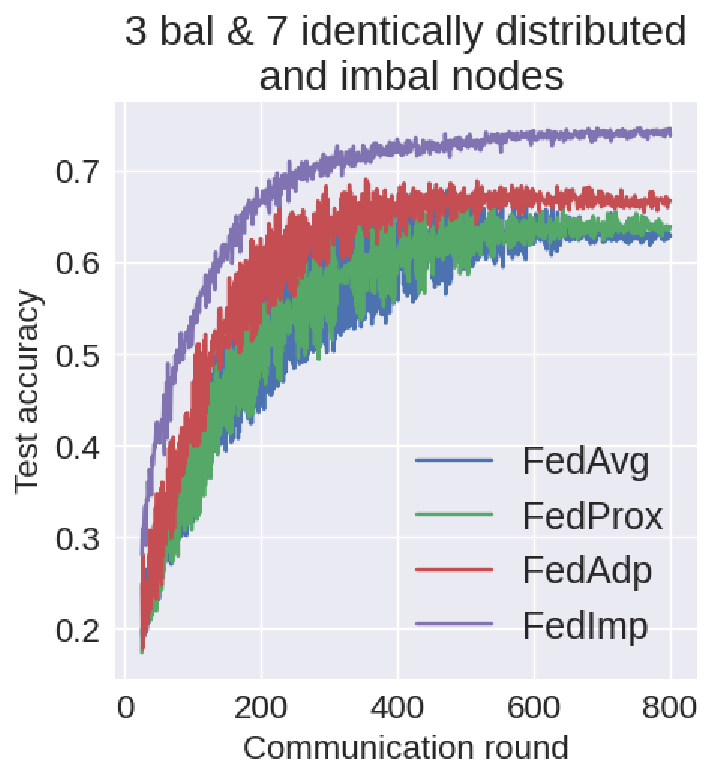}}
  \caption{\textbf{4-layer CNN} model for \textbf{CIFAR-10} results on test accuracy over communication rounds of FedAvg, FedProx, FedAdp, and FedImp with different levels of heterogeneous data distribution over participating nodes.} 
  \vspace{-10pt}
  \label{fig:cifar_2}
\end{figure}

\textit{4-Layer CNN for CIFAR-10 results:} Figure \ref{fig:cifar_2} illustrates test accuracy across communication rounds for FedAvg, FedProx, FedAdp, and FedImp, considering various scenarios of heterogeneous data distribution among nodes using the 4-layer CNN model. In scenarios involving only 3 imbalanced data nodes, there is not a notable disparity in algorithm performance. However, in scenarios with a greater number of imbalanced data nodes, both FedImp and FedAdp demonstrate a superior convergence rate compared to FedAvg and FedProx. Table \ref{tab:cifar_2} displays the number of communication rounds required to attain over the target accuracy on the test set using the 4-layer CNN model in different testing scenarios. 

\begin{itemize}
    \item In a setting featuring 7 balanced nodes and 3 imbalanced nodes, FedImp achieves an over 83\% accuracy after 533 rounds, while FedAvg and FedProx achieve similar target accuracy slower, at 577 and 677 rounds respectively. However, FedImp is slightly slower than FedAdp, which requires 528 rounds.
    \item Transitioning to the case of 5 balanced nodes and 5 imbalanced nodes, FedImp requires 296 rounds to surpass the 80\% accuracy target. FedImp has demonstrated a substantially lower number of communication rounds as opposed to 398 rounds for FedAdp, 526 rounds for FedAvg, and 529 rounds for FedProx. This signifies a decrease in the number of rounds by around 43.7\%, 44\%, and 25.6\% when comparing FedImp to FedAvg, FedProx, and FedAdp in the order given.
    \item In the scenario with 3 balanced nodes and 7 imbalanced nodes, while FedAvg and FedProx cannot reach the target accuracy of 80\%, FedImp demonstrates a slower convergence, reaching convergence in 599 rounds, whereas FedAdp achieves the same in 424 rounds.
    \item In the scenario of 3 balanced nodes + 7 identically distributed and imbalanced nodes, FedImp is still the only algorithm that can obtain 72\% or higher accuracy level. In contrast, FedProx, FedAdp, and FedImp converge at lower accuracy of around 60\% to 65\%.
\end{itemize} 

\begin{table}
\centering
\caption{The number of communication rounds for FL algorithms to reach over target test accuracy with the \textbf{4-layer CNN} model for \textbf{CIFAR-10} data. N/A indicates that the algorithm cannot achieve the target accuracy.}
\label{tab:cifar_2}
\begin{tabular}{|m{2cm}|c|c|c|c|c|} \hline
\textbf{Testing scenario}  &ACC & \textbf{FedAvg} & \textbf{FedProx} & \textbf{FedAdp} & \textbf{FedImp} \\ \hline
7 bal + 3 imbal & 83\% & 577 & 677 & 528 & 533 \\ %\hline
5 bal + 5 imbal & 80\% & 526 & 529 & 398 & 296 \\ %\hline
3 bal + 7 imbal & 80\% & N/A & N/A & 424 & 599 \\ %\hline
3 bal + 7 iden. dist. imbal & 72\% & N/A  & N/A & N/A & 310 \\ \hline
\end{tabular}
\vspace{-10pt}
\end{table}

\begin{table}[h]
\centering
\caption{Communication Overhead Comparison for Different FL Algorithms. The reported per-round overhead for FedImp (25.5 MB) includes the minor additional cost of transmitting the scalar weighting factor \( \psi_i \).}
\resizebox{1\linewidth}{!}{%
\begin{tabular}{|c|c|c|c|}
\hline
\textbf{Algorithm} & \textbf{Rounds} & \textbf{Com. overhead} & \textbf{Total com.} \\
& \textbf{to converse} & \textbf{per round (MB)} & \textbf{cost (MB)} \\
\hline
FedAvg  & 452 & 25.0 & 11300.0 \\
FedProx & 363 & 25.0 & 9075.0 \\
FedImp  & 252 & 25.5 & 6426.0 \\
\hline
\end{tabular}}
\label{tab:comm_overhead}
\end{table}

To assess the feasibility of FedImp in real-world FL deployments, we conducted additional experiments to measure its communication overhead in comparison to FedAvg and FedProx. Since communication efficiency is a critical factor in FL, particularly in IoT, mobile computing, and healthcare applications, understanding the bandwidth requirements of FedImp is essential. The objective was to analyze the following three aspects:

\begin{itemize}
    \item The number of communication rounds required to reach 80\% accuracy.
    \item The communication overhead per round (i.e., the size of model updates exchanged between clients and the server in MB).
    \item The total communication cost until convergence, computed as the product of per-round overhead and the number of rounds.
\end{itemize}

Each client transmits and receives the full model update in every round. The base model size used in this experiment was 2.5 MB, consistent across all algorithms. Since FedImp requires an additional impurity-based weighting step, there was a minor increase in communication overhead due to the transmission of the weighting factor $\psi_i$, though this is negligible compared to the model parameters. Table~\ref{tab:comm_overhead} presents the experimental results comparing FedAvg, FedProx, and FedImp in terms of communication efficiency. The slight increase in FedImp's per-round communication cost (25.5 MB) compared to FedAvg and FedProx (25.0 MB) accounts for the additional transmission of the scalar weighting factor \( \psi_i \). Although negligible in size, it is included in the reported overhead for completeness. 

As seen in the results, FedImp consistently requires fewer communication rounds to reach the same target accuracy compared to FedAvg and FedProx. Specifically, FedImp reduces the number of rounds by approximately 44.2\% compared to FedAvg and 30.6\% compared to FedProx. This results in a significant reduction in the total communication cost, with FedImp reducing bandwidth consumption by 43.1\% compared to FedAvg and 29.2\% compared to FedProx. Although FedImp introduces a slight increase in per-round communication overhead due to the transmission of impurity-based weights, this additional cost is negligible. The efficiency gain from faster convergence significantly outweighs this minor overhead, making FedImp a more communication-efficient choice in federated learning deployments.

\subsubsection{Discussion}

Based on the obtained results, we can observe that the proposed FedImp algorithm can achieve comparable or better test accuracy using considerably fewer communication rounds than FedAvg, FedProx, and FedAdp. This is particularly evident in cases with a higher number of imbalanced data nodes. These results demonstrate that FedImp offers a significant advantage in federated learning scenarios with non-IID data distributions. By requiring fewer communication rounds to achieve target accuracy, FedImp reduces communication overhead and improves the efficiency of the FL process. This is particularly important for large-scale deployments where communication costs can be substantial.

\paragraph{Why FedImp Outperforms FedAvg, FedProx, and FedAdp}

The superior performance of FedImp in reducing communication rounds stems from its entropy-based impurity weighting mechanism, which dynamically assigns aggregation weights based on the informational richness of each client’s local dataset. In contrast, traditional FL aggregation methods rely on either static or gradient-based weighting, which may not sufficiently address data heterogeneity. Below, we provide a detailed analysis of why FedImp achieves faster convergence.

\textbf{Adaptive Weighting for Improved Model Aggregation}: FedAvg applies uniform weighting based on dataset size, assuming that clients contribute equally to the global model. However, in non-IID settings, larger datasets may not necessarily contain diverse data, leading to slower generalization. FedImp, in contrast, assigns weights based on entropy. Higher entropy values indicate more diverse datasets, allowing FedImp to prioritize updates that contribute to better global model generalization.

\textbf{Mitigating the Impact of Skewed Data Contributions}: FedProx improves upon FedAvg by introducing a proximal term to constrain local updates. However, it does not differentiate between clients with high- and low-quality data distributions, which can still lead to inefficient aggregation. FedImp explicitly mitigates this issue by down-weighting nodes with highly imbalanced local datasets, ensuring that the global model is not disproportionately influenced by a small subset of labels.

\textbf{Faster Convergence via Balanced Information Contribution:} FedAdp attempts to improve convergence by assigning weights based on the local and global gradients. However, this approach struggles in extreme non-IID cases where clients lack overlapping label distributions, leading to misleading global gradient directions. FedImp instead leverages local dataset diversity rather than gradient similarity, enabling faster model stabilization even in highly heterogeneous scenarios.

\textbf{Empirical Evidence Supporting FedImp’s Faster Convergence:} The experimental results confirm that FedImp requires significantly fewer communication rounds to reach the same accuracy levels as FedAvg, FedProx, and FedAdp. By adaptively adjusting aggregation weights based on data impurity, FedImp consistently ensures that each communication round contributes maximally to generalization, thereby accelerating convergence while maintaining model robustness.

\paragraph{Impact of Temperature Parameter \(\tau\) in FedImp}

\begin{figure}[t]
    \centering
    \includegraphics[width=0.4\textwidth]{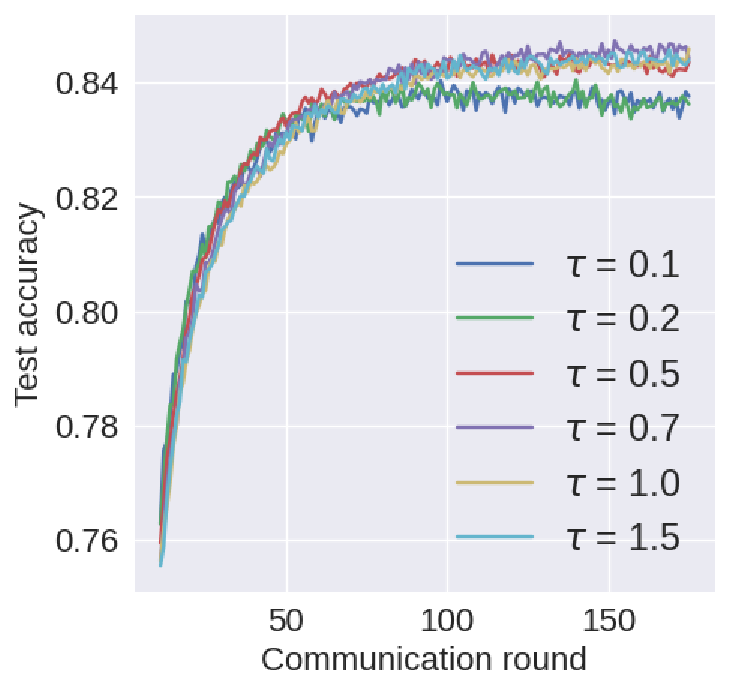}
    \caption{Test accuracy over communication rounds of FedImp with different \(\tau\). The data distribution setting is 5 balanced nodes + 5 imbalanced nodes with the EMNIST dataset and the MLP model.}
    \label{fig:temperature}
\end{figure}

Choosing the right temperature \(\tau\) for FedImp is crucial. Lower \(\tau\) values emphasize information-rich nodes while downplaying others, but small \(\tau\) values do not guarantee optimal performance. Figure \ref{fig:temperature} provides a visualization of test accuracy over communication rounds of FedImp with different \(\tau\) values in the scenario employing the EMNIST dataset and the MLP model with a data distribution setting of 5 balanced nodes and 5 imbalanced nodes. In this experimental setup, the most effective \(\tau\) is 0.7.

The value of \(\tau\) affects the weighting of the local models during aggregation, and choosing an inappropriate value may lead to suboptimal results. Further research is needed to develop automated methods or heuristics for selecting the optimal \(\tau\) value based on the characteristics of the dataset and the participating nodes. Despite this limitation, FedImp shows promising results in improving the efficiency and effectiveness of federated learning in scenarios with imbalanced data distributions.

\paragraph{Privacy and Security Considerations}
FedImp operates within the standard federated learning framework, where raw data remains on the client device, thereby preserving privacy by design. However, since aggregation weights are derived from local data statistics (entropy), it is important to consider potential privacy implications in adversarial environments.

Fortunately, the impurity score \( S_i \) and weighting factor \( \psi_i \) can be computed entirely on the client side and do not require sharing any intermediate statistics or raw data with the server. This makes FedImp compatible with common privacy-enhancing mechanisms such as Differential Privacy (DP) and Secure Aggregation (SA). For example, DP can be applied to local model updates before transmission, and SA protocols can be used to securely aggregate the weighted updates without revealing individual contributions.

Future work may explore combining FedImp with formal DP guarantees or integrating it into secure federated protocols, particularly in high-stakes domains like healthcare or finance where both convergence efficiency and privacy are paramount.

\paragraph{Toward Real-World Federated Settings}

While our experiments use controlled simulations with benchmark datasets (EMNIST and CIFAR-10) and synthetic non-IID partitions, we acknowledge the importance of validating FedImp in real-world federated learning deployments. Such environments introduce additional factors like device availability, communication delays, hardware heterogeneity, and natural data drift. Evaluating FedImp under these conditions—such as with real mobile users or healthcare data—remains a promising direction for future research to further assess its robustness and generalizability.

\paragraph{Potential Limitation of Entropy-Based Weighting}

While FedImp demonstrates strong performance across various non-IID scenarios, it has a limitation in the scenario where the entropy-based weighting is less effective. An exceptional scenario where clients are highly specialized (low entropy) and have mutually exclusive classes poses a serious challenge to FL in general as well as to FedImp. In this context, each client is often limited to only one or two classes, while there is no overlapping classes among the clients at all. Such a scenario will lead to a uniform aggregation weighting factor of the clients and reduce the effectiveness of the proposed algorithm. In such a case, combining entropy with other signals, such as inter-client label coverage or global class frequency awareness, could be a promising solution to address the challenge.

\section{Conclusion}
\label{sec:conclusion}
This paper presents FedImp, a novel FL algorithm for addressing the challenge of non-IID data. Our method entails assessing the contribution of each participating node by analyzing the informational entropy of its data. These contributions are subsequently standardized to produce unique weights for aggregating the global model. Through extensive experimental evaluations, FedImp showcases its superior convergence rate compared to FedAvg, FedProx, and FedAdp in various scenarios with non-IID data. Experimental findings demonstrate decreases of communication round up to 64.4\%, 27.8\%, and 66.7\% on the EMNIST dataset, relative to FedAvg, FedProx, and FedAdp, respectively. Moreover, on CIFAR-10 dataset, FedImp achieves great reductions, ranging up to 44.2\%, 44\%, and 25.6\% compared to FedAvg, FedProx, and FedAdp, respectively. In scenarios with a high number of imbalanced data nodes, such as the case of 3 balanced nodes and 7 identically distributed and imbalanced nodes, FedImp often emerges as the only algorithm among those evaluated that can reach the target convergence accuracy.

While FedImp offers advantages, practical FL deployment requires considering factors like device computational constraints and privacy implications. In future work, we plan to evaluate FedImp's performance across diverse models and datasets, and integrate it with privacy-preserving techniques, such as Differential Privacy and Secure Aggregation, to mitigate these risks. Robust aggregation or adversarial defenses could further strengthen FedImp, making it more suitable for real-world applications where both efficiency and privacy are essential. Another promising direction is to explore alternative impurity or diversity measures beyond Shannon entropy, such as Gini impurity, Simpson’s index, or kernel-based diversity metrics. These alternatives may capture different aspects of client data heterogeneity and could offer improved sensitivity or robustness in weighting decisions under specific distributional scenarios.

\bibliographystyle{IEEEtran}
\bibliography{reference}

@article{mammen2021federated,
  title={Federated learning: Opportunities and challenges},
  author={Mammen, Priyanka Mary},
  journal={arXiv preprint arXiv:2101.05428},
  year={2021}
}

@article{kairouz2021advances,
  title={Advances and open problems in federated learning},
  author={Kairouz, Peter and McMahan, H Brendan and Avent, Brendan and Bellet, Aur{\'e}lien and Bennis, Mehdi and Bhagoji, Arjun Nitin and Bonawitz, Kallista and Charles, Zachary and Cormode, Graham and Cummings, Rachel and others},
  journal={Foundations and Trends{\textregistered} in Machine Learning},
  volume={14},
  number={1--2},
  pages={1--210},
  year={2021},
  publisher={Now Publishers, Inc.}
}

@article{zhu2021federated,
  title={Federated learning on non-IID data: A survey},
  author={Zhu, Hangyu and Xu, Jinjin and Liu, Shiqing and Jin, Yaochu},
  journal={Neurocomputing},
  volume={465},
  pages={371--390},
  year={2021},
  publisher={Elsevier}
}

@article{liu2020systematic,
  title={A systematic literature review on federated learning: From a model quality perspective},
  author={Liu, Yi and Zhang, Li and Ge, Ning and Li, Guanghao},
  journal={arXiv preprint arXiv:2012.01973},
  year={2020}
}

@inproceedings{mcmahan2017communication,
  title={Communication-efficient learning of deep networks from decentralized data},
  author={McMahan, Brendan and Moore, Eider and Ramage, Daniel and Hampson, Seth and y Arcas, Blaise Aguera},
  booktitle={Artificial intelligence and statistics},
  pages={1273--1282},
  year={2017},
  organization={PMLR}
}

@article{wu2021fast,
  title={Fast-convergent federated learning with adaptive weighting},
  author={Wu, Hongda and Wang, Ping},
  journal={IEEE Transactions on Cognitive Communications and Networking},
  volume={7},
  number={4},
  pages={1078--1088},
  year={2021},
  publisher={IEEE}
}

@article{hsu2019measuring,
  title={Measuring the effects of non-identical data distribution for federated visual classification},
  author={Hsu, Tzu-Ming Harry and Qi, Hang and Brown, Matthew},
  journal={arXiv preprint arXiv:1909.06335},
  year={2019}
}

@inproceedings{yeganeh2020inverse,
  title={Inverse distance aggregation for federated learning with non-iid data},
  author={Yeganeh, Yousef and Farshad, Azade and Navab, Nassir and Albarqouni, Shadi},
  booktitle={Domain Adaptation and Representation Transfer, and Distributed and Collaborative Learning: Second MICCAI Workshop, DART 2020, and First MICCAI Workshop, DCL 2020, Held in Conjunction with MICCAI 2020, Lima, Peru, October 4--8, 2020, Proceedings 2},
  pages={150--159},
  year={2020},
  organization={Springer}
}

@article{acar2021federated,
  title={Federated learning based on dynamic regularization},
  author={Acar, Durmus Alp Emre and Zhao, Yue and Navarro, Ramon Matas and Mattina, Matthew and Whatmough, Paul N and Saligrama, Venkatesh},
  journal={arXiv preprint arXiv:2111.04263},
  year={2021}
}

@inproceedings{vahidian2021personalized,
  title={Personalized federated learning by structured and unstructured pruning under data heterogeneity},
  author={Vahidian, Saeed and Morafah, Mahdi and Lin, Bill},
  booktitle={2021 IEEE 41st international conference on distributed computing systems workshops (ICDCSW)},
  pages={27--34},
  year={2021},
  organization={IEEE}
}

@TECHREPORT{Krizhevsky09learningmultiple,
            author={Alex Krizhevsky},
            title={Learning multiple layers of features from tiny images},
            institution={},
            year={2009}
}

@article{deng2012mnist,
  title={The mnist database of handwritten digit images for machine learning research [best of the web]},
  author={Deng, Li},
  journal={IEEE signal processing magazine},
  volume={29},
  number={6},
  pages={141--142},
  year={2012},
  publisher={IEEE}
}

@article{reddi2020adaptive,
  title={Adaptive federated optimization},
  author={Reddi, Sashank and Charles, Zachary and Zaheer, Manzil and Garrett, Zachary and Rush, Keith and Kone{\v{c}}n{\`y}, Jakub and Kumar, Sanjiv and McMahan, H Brendan},
  journal={arXiv preprint arXiv:2003.00295},
  year={2020}
}

@article{li2020federated,
  title={Federated optimization in heterogeneous networks},
  author={Li, Tian and Sahu, Anit Kumar and Zaheer, Manzil and Sanjabi, Maziar and Talwalkar, Ameet and Smith, Virginia},
  journal={Proceedings of Machine learning and systems},
  volume={2},
  pages={429--450},
  year={2020}
}

@article{fotohi2024lightweight,
  title={A lightweight and secure deep learning model for privacy-preserving federated learning in intelligent enterprises},
  author={Fotohi, Reza and Aliee, Fereidoon Shams and Farahani, Bahar},
  journal={IEEE Internet of Things Journal},
  year={2024},
  publisher={IEEE}
}

@article{fotohi2024decentralized,
  title={Decentralized and robust privacy-preserving model using blockchain-enabled federated deep learning in intelligent enterprises},
  author={Fotohi, Reza and Aliee, Fereidoon Shams and Farahani, Bahar},
  journal={Applied Soft Computing},
  volume={161},
  pages={111764},
  year={2024},
  publisher={Elsevier}
}
%\begin{thebibliography}{00}

\end{document}